\documentclass[british,10pt,twocolumn,letterpaper]{article}
\usepackage[T1]{fontenc}
\usepackage[latin9]{luainputenc}
\usepackage{array}
\usepackage{rotating}
\usepackage{bbding}
\usepackage{multirow}
\usepackage{amsmath}
\usepackage{amssymb}
\usepackage{graphicx}

\makeatletter

\providecommand{\tabularnewline}{\\}
\newcommand{\lyxdot}{.}

\usepackage{cvpr}
\usepackage{times}
\usepackage{epsfig}
\usepackage{graphicx}
\usepackage{amsmath}
\usepackage{amstext}

\usepackage{placeins}
\usepackage{color}
\usepackage{colortbl}
\usepackage{rotating}
\definecolor{lightgray}{gray}{0.8}
\definecolor{verylightgray}{gray}{0.9}

\usepackage[pagebackref=true,breaklinks=true,letterpaper=true,colorlinks,bookmarks=false]{hyperref}

\cvprfinalcopy
\def\cvprPaperID{1802} 

\ifcvprfinal\fi

\@ifundefined{showcaptionsetup}{}{%
 \PassOptionsToPackage{caption=false}{subfig}}
\usepackage{subfig}
\makeatother

\usepackage{babel}
\begin{document}
\makeatletter  
\renewcommand{\paragraph}{%
\@startsection{paragraph}{4}%
 {\z@}{0.5ex \@plus 1ex \@minus .2ex}{-0.5em}%
  {\normalfont \normalsize \bfseries}%
} 

\let\originalparagraph\paragraph 
\renewcommand{\paragraph}[2][.]{\originalparagraph{#2#1}}

\makeatother

\title{\vspace{-1em}
Exploiting Saliency for Object Segmentation from Image Level Labels}

\author{\vspace{0em}
\setlength\tabcolsep{1em}
\begin{tabular}{ccc} 
Seong Joon Oh\footnotemark[2] & Rodrigo Benenson\footnotemark[2] & Anna   Khoreva\footnotemark[2] \tabularnewline
\texttt{\small{}joon@mpi-inf.mpg.de} & \texttt{\small{}benenson@mpi-inf.mpg.de} & \texttt{\small{}khoreva@mpi-inf.mpg.de} \tabularnewline
\vspace{-0.8em} & & \tabularnewline
Zeynep Akata\footnotemark[2]\hspace{0.35em}\textsuperscript{,}\footnotemark[3] & Mario Fritz\footnotemark[2] & Bernt Schiele\footnotemark[2] \tabularnewline
\texttt{\small{}Z.Akata@uva.nl} & \texttt{\small{}mfritz@mpi-inf.mpg.de} & \texttt{\small{}schiele@mpi-inf.mpg.de} \tabularnewline
\end{tabular}
\\
\\
\renewcommand{\arraystretch}{0.9}
\begin{tabular}{ccc} 
\footnotemark[2]\enskip{}\normalsize{Max Planck Institute for Informatics}  & & \footnotemark[3]\enskip{}\normalsize{Amsterdam Machine Learning Lab} \tabularnewline
\normalsize{Saarland Informatics Campus} & & \normalsize{University of Amsterdam} \tabularnewline
\normalsize{Saarbr\"ucken, Germany} & & \normalsize{Amsterdam, the Netherlands} \tabularnewline
\end{tabular}
}
\maketitle



\begin{abstract}
There have been remarkable improvements in the semantic labelling
task in the recent years. However, the state of the art methods rely
on large-scale pixel-level annotations. This paper studies the problem
of training a pixel-wise semantic labeller network from image-level
annotations of the present object classes. Recently, it has been shown
that high quality seeds indicating discriminative object regions can
be obtained from image-level labels. Without additional information,
obtaining the full extent of the object is an inherently ill-posed
problem due to co-occurrences. We propose using a saliency model as
additional information and hereby exploit prior knowledge on the
object extent and image statistics. We show how to combine both information
sources in order to recover $80\%$ of the fully supervised performance
\textendash{} which is the new state of the art in weakly supervised
training for pixel-wise semantic labelling. The code is available at \url{https://goo.gl/KygSeb}.

\end{abstract}

\section{\label{sec:Introduction}Introduction}

\noindent Semantic image labelling provides rich information about
scenes, but comes at the cost of requiring pixel-wise labelled training
data. The accuracy of convnet-based models correlates strongly with
the amount of available training data. Collection and annotation of
data have become a bottleneck for progress. This problem has raised
interest in exploring partially supervised data or different means
of supervision, which represents different tradeoffs between annotation
efforts and yields in terms of supervision signal for the learning
task. For tasks like semantic segmentation there is a need to investigate
the minimal supervision to reach the quality comparable to the fully
supervised case.

A reasonable starting point considers that all training images have
image-level labels to indicate the presence or absence of the classes
of interest. The weakly supervised learning problem can be seen as
a specific instance of learning from constraints \cite{Shcherbatyi2016Gcpr,Xu2015CvprWeakSegmentation}.
Instead of explicitly supervising the output, the available labels
provide a constraint on the desired output. If an image label is absent,
no pixel in the image should take that label; if an image label is
present at least in one pixel the image must take that label. However,
the objects of interest are rarely single pixel. Thus to enforce larger
output regions size, shape, or appearance priors are commonly employed
(either explicitly or implicitly). 

\begin{figure}
\begin{centering}
\includegraphics[width=0.9\columnwidth,height=0.4\columnwidth]{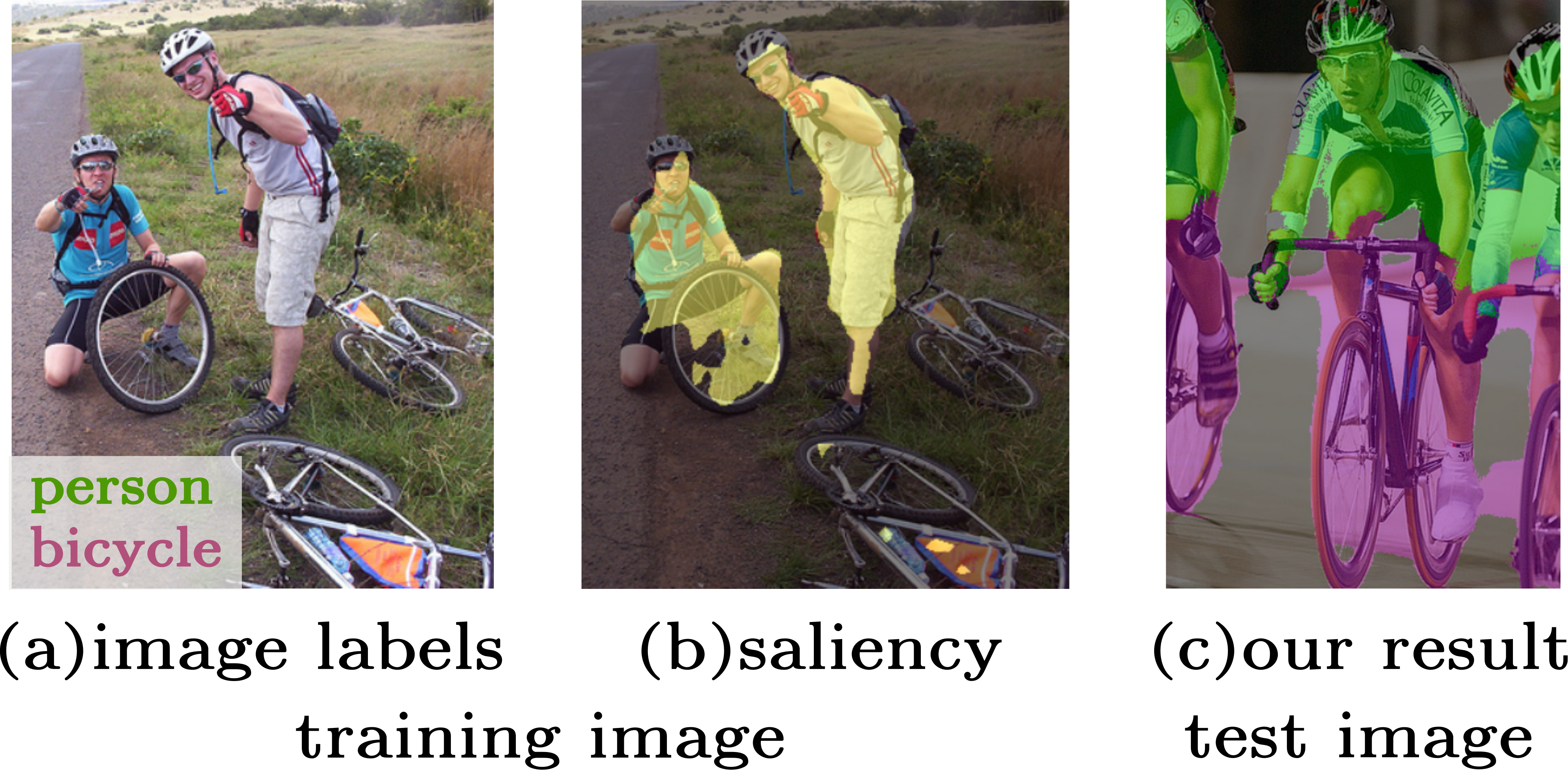}
\par\end{centering}
\caption{\label{fig:teaser}We train a semantic labelling network with (a)
image-level labels and (b) saliency masks, to generate (c) a pixel-wise
labelling of object classes at test time.}
\end{figure}

Another reason for exploiting priors, is the fact that the task is
fundamentally ambiguous. Strongly co-occurring categories (such as
train and rails, sculls and oars, snow-bikes and snow) cannot be separated
without additional information. Because additional information is
needed to solve the task, previous work have explored different avenues,
including class-specific size priors \cite{Pathak2015Iccv}, crawling
additional images \cite{Pinheiro2015Cvpr,Wei2015ArXiv}, or requesting
corrections from a human judge \cite{Kolesnikov2016Bmvc,Saleh2016Eccv}.

Despite these efforts, the quality of the current best results on
the task seems to level out at $\sim\negmedspace75\%$ of the fully
supervised case. Therefore, we argue that additional information sources
have to be explored to complement the image level label supervision
\textendash{} in particular addressing the inherent ambiguities of
the task. In this work, we propose to exploit class-agnostic saliency
as a new ingredient to train for class-specific pixel labelling; and
show new state of the art results on Pascal VOC 2012 semantic labelling
with image label supervision.

We decompose the problem of object segmentation from image labels
into two separate ones: finding the object location (any point on
the object), and finding the object's extent. Finding the object extent
can be equivalently seen as finding the background area in an image. 

For object location we exploit the fact that image classifiers are
sensitive to the discriminative areas of an image. Thus training using
the image labels enables to find high confidence points over the objects
classes of interest (we call these ``object seeds''), as well as
high confidence regions for background. A classifier however will
struggle to delineate the fine details of an object instance, since
these might not be particularly discriminative.

For finding the object extent, we exploit the fact that a large portion
of photos aim at capturing a subject. Using class-agnostic object
saliency we can find the segment corresponding to some of the detected
object seeds. Albeit saliency is noisy, it provides information delineating
the object extent beyond what seeds can indicate. Our experiment show
that this is an effective source of additional information. Our saliency
model is itself trained from bounding box annotations only. At no
point of our pipeline accurate pixel-wise annotations are used.

In this paper we provide an analysis of the factors that influence
the seeds generation, explore the utility of saliency for the task,
and report best known results both when using image labels only and
image labels with additional data. In summary, our contributions are: 

\begin{itemize}
\item Propose an effective method for combining seed and saliency for weakly supervised semantic segmentation. Our method achieves the best performance among
the known works that utilise image level supervision with or without
additional external data.
\item Compare recent seed methods side by side, and analyse the
    importance of saliency towards the final quality.
\end{itemize}

\S \ref{sec:architecture} presents our overall architecture,
\S \ref{sec:goods-seeds} investigates suitable object seeds,
and \S \ref{sec:object-extent} describes how we use saliency
to guide the convnet training. Finally \S \ref{sec:Experiments}
discusses the experimental setup, and presents our key results.

\section{\label{sec:Related-work}Related work}

\noindent
The last years have seen a renewed interest on weakly supervised training.
For semantic labelling, different forms of supervision have been explored:
image labels \cite{Pathak2015Iclrw,Pathak2015Iccv,Papandreou2015Iccv,Pinheiro2015Cvpr,Wei2015ArXiv,kolesnikov2016seed},
points \cite{Bearman2015ArXiv}, scribbles \cite{Xu2015CvprWeakSegmentation,Lin2016CvprScribbleSup},
and bounding boxes \cite{Dai2015Iccv,Papandreou2015Iccv,Khoreva2016Arxiv}.
In this work we focus on image labels as the main form of supervision.

\paragraph{Object seeds}

Multiple works have considered using a trained classifier (from image
level labels) to find areas of the image that belong to a given class,
without necessarily enforcing to cover the full object extent (high
precision, low recall). Starting from simple strategies such as ``probing
classifier with different image areas occluded'' \cite{Zeiler2014Eccv},
or back-propagating the class score gradient on the image \cite{Simonyan2014Iclr};
significantly more involved strategies have been proposed, mainly
by modifying the back-propagation strategy \cite{Springenberg2015Iclrw,Zhang2016Eccv,Shimoda2016Eccv},
or by solving a per-image optimization problem \cite{Cao2015Iccv}.
All these strategies provide some degree of empirical success but
lack a clear theoretical justification, and tend to have rather noisy
outputs.\\
Another approach considers modifying the classifier training procedure
so as to have it generate object masks as by-product of a forward-pass.
This can be achieved by adding a global a max-pooling \cite{Pinheiro2015Cvpr}
or mean-pooling layer \cite{zhou2015cnnlocalization} in the last
stages of the classifier.\\
In this work we provide an empirical comparison of existing seeders,
and explore variants of the mean-pooling approach \cite{zhou2015cnnlocalization}
(\S \ref{sec:goods-seeds}).

\paragraph{Pixel labelling from image level supervision}

Initial work approached this problem by adapting multiple-instance
learning \cite{Pathak2015Iclrw} and expectation-maximization techniques
\cite{Papandreou2015Iccv}, to the semantic labelling case. Without
additional priors only poor results are obtained. Using superpixels
to inform about the object shape helps \cite{Pinheiro2015Cvpr,Xu2015CvprWeakSegmentation}
and so does using priors on the object size \cite{Pathak2015Iccv}.
\cite{kolesnikov2016seed} carefully uses CRFs to propagate the seeds
across the image during training, while \cite{Qi2016Eccv} exploits
segment proposals for this.\\
Most methods compared propose each a new procedure to train a semantic
labelling convnet. One exception is \cite{Shimoda2016Eccv} which
fuses at test time guided back-propagation \cite{Springenberg2015Iclrw}
at multiple convnet layers to generate class-wise heatmaps. They do
this over a convnet trained for classification. Being based on classifier,
their output masks only partially capture the object extents, as reflected
in the comparatively low performance (table \ref{tab:pascal-results-others}).\\
Recognizing the ill-posed nature of the problem, \cite{Kolesnikov2016Bmvc}
and \cite{Saleh2016Eccv} propose to collect user-feedback as additional
information to guide the training of a segmentation convnet.\\
The closest work to our approach is \cite{Wei2015ArXiv}, which also
uses saliency as a cue to improve weakly supervised semantic segmentation.
There are however a number of differences. First, they use a curriculum
learning to expose the segmentation convnet first with simple images,
and later with more complex ones. We do not need such curriculum,
yet reach better results. Second, they use a manually crafted class-agnostic
saliency method, while we use a deep learning based one (which provides
better cues). Third, their training procedure uses $\sim\negthinspace40\text{k}$
additional images of the classes of interest crawled from the web;
we do not use such class-specific external data. Fourth, we report
significantly better results, showing in better light the potential
of saliency as additional information to guide weakly supervised semantic
object labelling. \\
The seminal work \cite{Vezhnevets2011Iccv} proposed to use
``objectness'' map from bounding boxes to guide the semantic
segmentation. By using bounding boxes, these maps end up being
diffuse; in contrast, our saliency map has sharp object
boundaries, thus giving more precise guidance to the semantic
labeller.

\paragraph{Detection boxes from image level supervision}

Detecting object boxes from image labels has similar challenges as
pixel labelling. The object location and extent need to be found.
State of the art techniques for this task \cite{Bilen2016Cvpr,Teh2016Bmvc,Kantorov2016Eccv}
learn to re-score detection proposals using two stream architectures
that once trained separate ``objectness'' scores from class scores.
These architecture echo with our approach, where the seeds provide
information about the class scores at each pixel (albeit with low
recall for foreground classes), and the saliency output provides a
per-pixel (class agnostic) ``objectness'' score.

\paragraph{Saliency}

Image saliency has multiple connotations, it can refer to a spatial
probability map of where a person might look first \cite{Yamada2010Accv},
a probability map of which object a person might look first \cite{Li2014CvprSaliency},
or a binary mask segmenting the one object a person is most likely
to look first \cite{Borji2015Tip,Shi2016Pami}. We employ the last
definition in this paper. Note that this notion is class-agnostic,
and refers more to the composition of the image, than the specific object
category.\\
Like most computer vision areas, hand-crafted methods \cite{Jiang2013Cvpr,Margolin2013Cvpr,Cheng2015Pami}
have now been surpassed by convnet based approaches \cite{Zhao2015Cvpr,Li2016Tip,Li2016Cvpr}
for object saliency. In this paper we use saliency as an ingredient:
improved saliency models would lead to improved results for our method.
We describe in \S \ref{sec:Saliency-details} our saliency model
design, trained itself in a weakly supervised fashion from bounding
boxes.

\paragraph{Semantic labelling}

Even when pixel-level annotations are provided (fully supervised case),
the task of semantic labelling is far from being solved. Multiple convnet
architectures have been proposed, including recurrent networks \cite{Pinheiro2014Icml},
encoder-decoders \cite{Noh2015Iccv,Badrinarayanan2015ArxivSegNet},
up-sampling layers \cite{Long2015Cvpr}, using skip layers \cite{Bansal2016Arxiv},
or dilated convolutions \cite{Chen2016ArxivDeeplabv2,Yu2016Iclr},
to name a few. Most of them build upon classification architectures
such as VGG \cite{Simonyan2015Iclr} or ResNet \cite{He2016Cvpr}.
For comparison with previous work, our experiments are based on the
popular DeepLab \cite{Chen2016ArxivDeeplabv2} architecture.

\section{\label{sec:architecture} Guided Segmentation architecture}

While previous work has explored sophisticated training losses
or involved pipelines, we focus on saliency as an effective
prior knowledge, and thus keep our architecture simple.

\begin{figure}
\begin{centering}
\includegraphics[width=0.95\columnwidth]{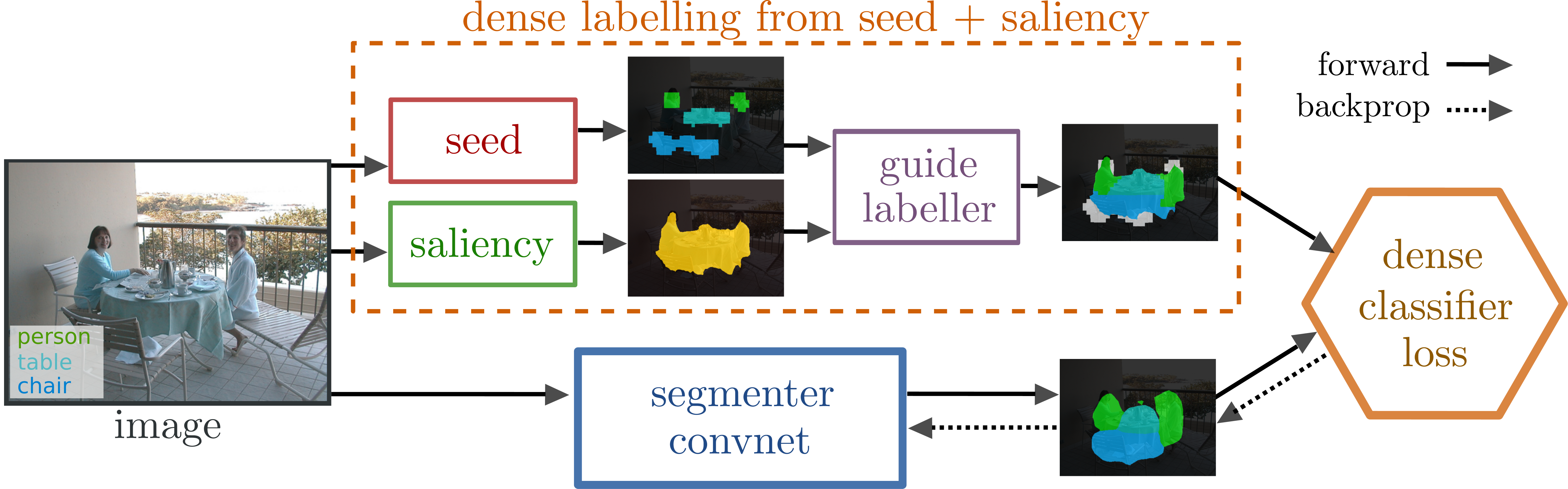}
\par\end{centering}
\caption{\label{fig:high-level-architecture}High level Guided Segmentation
architecture. }
\end{figure}

We approach the image-level supervised semantic segmentation problem
via a system with two modules (see figure \ref{fig:high-level-architecture}),
we name this architecture ``Guided Segmentation''. Given an image
and image-level labels, the ``guide labeller'' module combines cues
from a seeder (\S\ref{sec:goods-seeds}) and saliency (\S\ref{sec:object-extent})
sub-modules, producing a rough segmentation mask (the ``guide'').
Then a segmenter convnet is trained using the produced guide mask
as supervision. In this architecture the segmentation convnet is trained
in a fully-supervised procedure, using per pixel softmax
cross-entropy loss. 

In \S \ref{sec:goods-seeds} and \ref{sec:object-extent} we
explain how we build our guide labeller, by first generating seeds
(discriminative areas of objects of interest), and then extending
them to better cover the full object extents. 

\section{\label{sec:goods-seeds}Finding goods seeds}

\noindent There has been a recent burst of techniques for localising
objects from a classifier. Some approaches rely on image gradients
from a trained classifier \cite{Simonyan2014Iclr,Springenberg2015Iclrw,Zhang2016Eccv},
while the others propose to train global average pooling (GAP) based
 classifiers \cite{zhou2015cnnlocalization}. Although the classifier
 based localisation approach has a theoretical limitation that the
 training objective (image classification) does not match final goal
 (object locations), they have proved to be effective in practice.
 
 In this section, we review the seeder techniques side by side and
 compare their empirical performances. We report empirical results
 on different GAP architectures \cite{zhou2015cnnlocalization,kolesnikov2016seed,Chen2016ArxivDeeplabv2}.

\subsection{GAP}

\noindent GAP, or global average pooling layer, can be inserted in
the last or penultimate layer of a fully convolutional architecture,
which produces a dense prediction, to turn it into a classifier. The
resulting architecture is then trained with a classification loss,
and at test time the activation maps before the global average pooling
layer have been shown to contain localisation information \cite{zhou2015cnnlocaliz     ation}. 

In our analysis, we consider four different fully convolutional architectures
with a GAP layer: $\mathtt{GAP}\text{-}\mathtt{LowRes}$, $\mathtt{GAP}\text{-}\mathtt{HighRes}$,
$\mathtt{GAP}\text{-}\mathtt{DeepLab}$, and $\mathtt{GAP}\text{-}\mathtt{ROI}$.
The architectural differences are summarised in table \ref{tab:GAP-architectures},
 and the full details are provided in the supplementary materials.
 $\mathtt{GAP}\text{-}\mathtt{LowRes}$ \cite{zhou2015cnnlocalization}
 is essentially a fully convolutional version of VGG-16 \cite{Simonyan2015Iclr}.
 $\mathtt{GAP}\text{-}\mathtt{HighRes}$, inspired by \cite{kolesnikov2016seed},
 has $2$ times higher output resolution than $\mathtt{GAP}\text{-}\mathtt{LowRes}$.
 $\mathtt{GAP}\text{-}\mathtt{DeepLab}$ is a state of the art semantic
 segmenter DeepLab with a GAP layer over the dense score output. The
 main difference between $\mathtt{GAP}\text{-}\mathtt{HighRes}$ and
 $\mathtt{GAP}\text{-}\mathtt{DeepLab}$ is the presence of dilated
 convolutions. $\mathtt{GAP}\text{-}\mathtt{ROI}$ is a variant of
 $\mathtt{GAP}\text{-}\mathtt{HighRes}$ where we use the region of
 interest pooling to replace the sliding window convolutions in the
 last layers of VGG-16. $\mathtt{GAP}\text{-}\mathtt{ROI}$ is identical
 to $\mathtt{GAP}\text{-}\mathtt{HighRes}$, except for a slight structural
 variation.

\begin{figure}
\begin{centering}
\subfloat[Foreground categories]{\begin{centering}
\includegraphics[width=0.75\columnwidth]{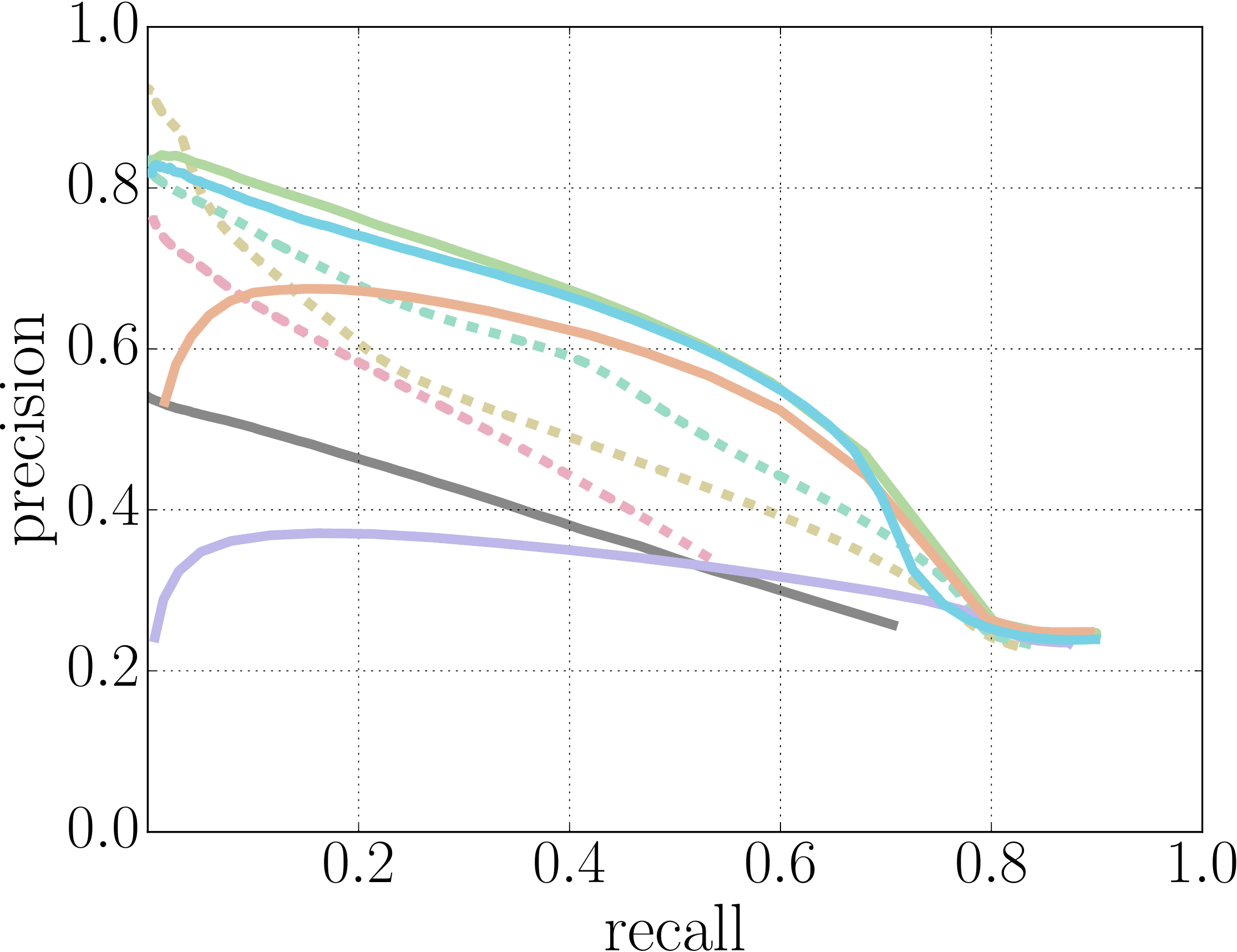}
\par\end{centering}
}
\par\end{centering}
\begin{centering}
\subfloat[Background category]{\begin{centering}
\includegraphics[width=0.75\columnwidth]{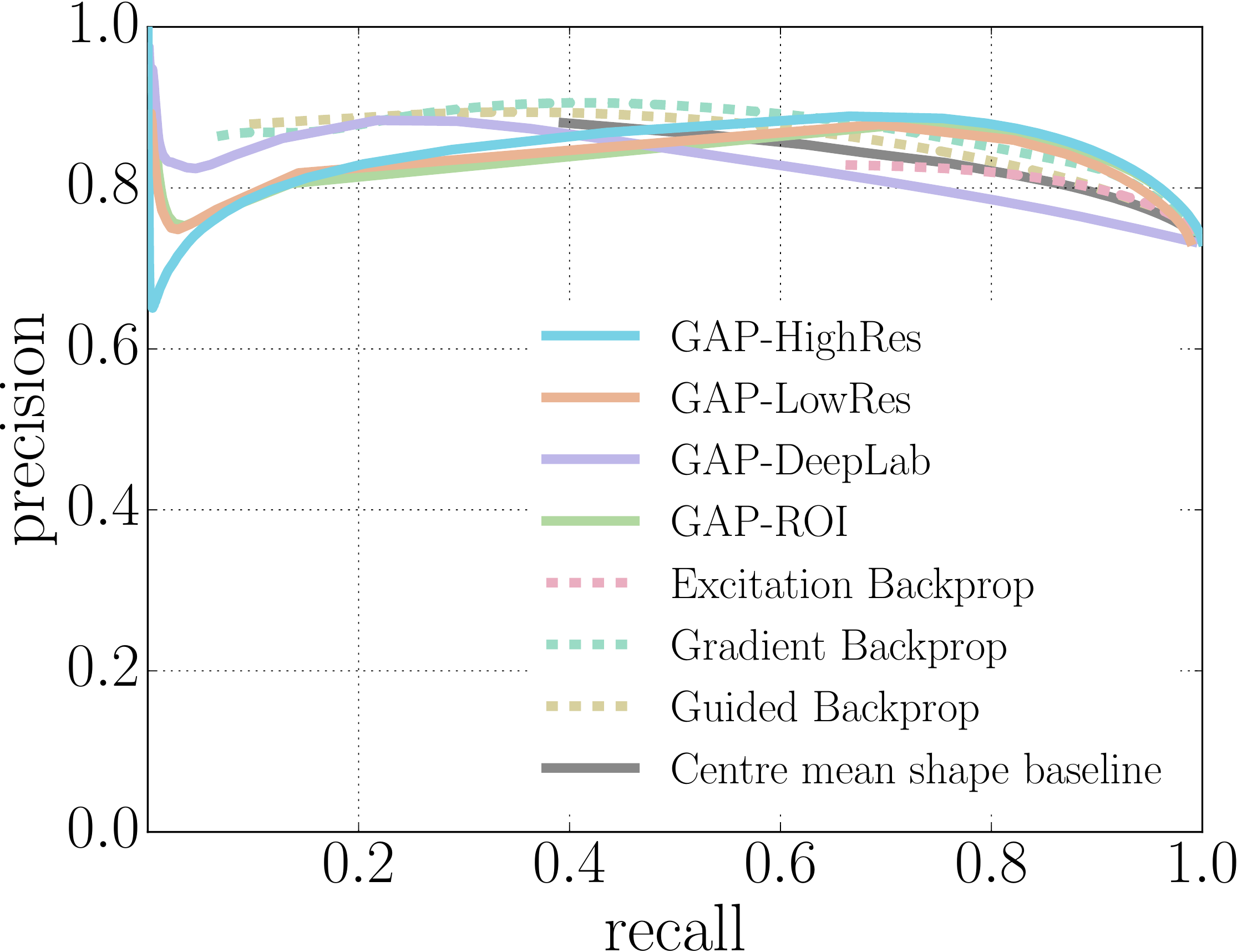}
\par\end{centering}
}
\par\end{centering}
\caption{\label{fig:comparing-seeds-methods}Precision-recall curves for different seeds. Foregrounds curves show the average precision and recall of the 20 foreground classes.}
\end{figure}

\begin{table}
\begin{centering}
\begin{tabular}{ccccc}
$\mathtt{GAP}$  & $\text{-}\mathtt{LowRes}$  & $\text{-}\mathtt{HighRes}$  & $\text{-}\mathtt{ROI}$  & $\text{-}\mathtt{DeepLab}$ \tabularnewline
 & \cite{zhou2015cnnlocalization} & \cite{kolesnikov2016seed} &  & \cite{Chen2016ArxivDeeplabv2}\tabularnewline
\hline 
high res. & \textbf{\scriptsize{}\XSolidBrush{}} & $\checkmark$ & $\checkmark$ & $\checkmark$\tabularnewline
dil. conv. & \textbf{\scriptsize{}\XSolidBrush{}} & \textbf{\scriptsize{}\XSolidBrush{}} & \textbf{\scriptsize{}\XSolidBrush{}} & $\checkmark$\tabularnewline
ROI pool & \textbf{\scriptsize{}\XSolidBrush{}} & \textbf{\scriptsize{}\XSolidBrush{}} & $\checkmark$ & \textbf{\scriptsize{}\XSolidBrush{}}\tabularnewline
\hline 
mP & 76.5 & 80.7 & 80.8 & 57.7\tabularnewline
mAP & 88.0 & 87.0 & 87.2 & 92.7\tabularnewline
\end{tabular}
\par\end{centering}

 \caption{\label{tab:GAP-architectures}Architectural comparisons among GAP
  variants together with classification (mAP) and localisation (mP;
  see text for details) performances. We compare the output resolution
  (high res.), use of the dilated convolutions (dil. conv.), and the
  region of interest pooling (ROI pool).}
\end{table}

\subsection{Empirical study}

 \noindent In this section, we empirically compare the seed methods
 side by side focusing on their utility for the final semantic segmentation
 task. Together with GAP methods discussed in the previous section,
 we consider the back-propagation family: Vanilla, Guided, and Excitation
 back-propagations \cite{Simonyan2014Iclr,Springenberg2015Iclrw,Zhang2016Eccv}.
 We include the centre mean shape baseline that always outputs the
 average mask shape; it works as a lower bound on the localisation
 performance.
 
 \paragraph{Evaluation}
 
 We evaluate each method on the \emph{val} set of the Pascal VOC 2012
 \cite{pascal-voc-2012} segmentation benchmark. We plot the foreground
 and background precision-recall curves in figure \ref{fig:comparing-seeds-methods}     .
 In the foreground case, we compute the mean precision and recall over
 the $20$ Pascal categories.
 
 We define mean precision (mP) as a summary metric for localisation
 performance. It averages the foreground precision at $20\%$ recall
 and the background precision at $80\%$ recall; $\mathrm{mP}=\frac{\mathrm{Prec}_{\mathrm{Fg}@20\%}+\mathrm{Prec}_{\mathrm{Bg}@80\%}}{2}$.
 Intuitively, for the foreground region we only need a small discriminative
 region, as saliency will fill in the extent; we thus care about precision
 at $\sim\negthinspace20\%$ recall. On the other hand, background
 has more diverse appearance and usually takes a larger region; we
 thus care about precision at $\sim\negthinspace80\%$ recall. Since
 we care about both, we take the average (as for the mAP metric). This
 metric has shown a good correlation with the final performance in
 our preliminary experiments.
 
 We measure the classification performance in the standard mean average
 precision (mAP) metric. 
 
 \paragraph{Implementation details}
 
 We train all four GAP network variants for multi-label image classification
 over the \emph{trainaug} set of Pascal VOC 2012. Full convnet training
 details are in the supplementary materials. At test time, we take
 the output per-class heatmaps before the GAP layer and normalise them
 by the maximal per-class scores. 
 
 For the back-propagation based methods, we obtain image (pseudo-)gradients
 from the VGG-16 \cite{Simonyan2015Iclr} classifier trained on the
 \emph{trainaug} set of Pascal VOC 2012 ($10\,582$ images in total).
 We take the maximal absolute gradient value across the RGB channels
 to generate a rough object mask (following \cite{Simonyan2014Iclr});
 it is successively smoothed first with vanilla Gaussian kernel and
 then with dense CRF \cite{Kraehenbuehl2011Nips}. 
 
 In both GAP and backprop variants, we mark pixels with all foreground
 class scores below $\tau$ as background; other pixels are marked
 according to the argmax foreground class.
 
 \paragraph{Results}
 
 See figure \ref{fig:comparing-seeds-methods} for the precision-recall
 curves. GAP variants have overall greater precision than backprop
 variants at the same recall rate. We note that the Guided backprop
 gives highest precision at a very low recall regime ($\sim\negthinspace5\%$),
 but the recall is too low to be useful. Among the GAP methods, $\mathtt{GAP}\text{     -}\mathtt{HighRes}$
 and $\mathtt{GAP}\text{-}\mathtt{ROI}$ give higher precisions over
 a wide range of recall. $\mathtt{GAP}\text{-}\mathtt{DeepLab}$ shows
 a significantly lower quality than any other GAP variants.

 \paragraph{Network matters for GAP}

 Table \ref{tab:GAP-architectures} shows detailed architectural comparisons
 and classification/localisation performances of the GAP variants.
 We observe that the network with higher resolution output has better
 localisation performance (80.7 mP for $\mathtt{GAP}\text{-}\mathtt{HighRes}$
 versus 76.5 mP for $\mathtt{GAP}\text{-}\mathtt{LowRes}$). Dilated
 convolutions significantly hurt the GAP performance (87.0 mP for $\mathtt{GAP}\text{-}\mathtt{HighRes}$
 versus 57.7 mP for $\mathtt{GAP}\text{-}\mathtt{DeepLab}$).  The
 architectural choice matters a lot for the localisation performance.
 This contrasts with the classification performances (mAP), which are
 stable across design choices. Intriguingly, $\mathtt{GAP}\text{-}\mathtt{DeepLab}$
 is in fact the best classifier and the worst seeder at the same time;
 better design choices for classifiers do not lead to better seeders. 
 
 We use $\mathtt{GAP}\text{-}\mathtt{HighRes}$ as the seeder module
 in the next sections. In \cite{kolesnikov2016seed}, foreground and
 background seeds are handled via different mechanisms; in our experiments
 we treat all the non-foreground region as background.

\section{\label{sec:object-extent}Finding the object extent}

Having generated a set of seeds indicating discriminative object areas,
the guide labeller needs to find the extent of the object instances
(\S\ref{sec:architecture}). 

Without any prior knowledge, it is very hard, if not impossible, to
learn the extent of objects only from images and image-level labels.
Image-level labels only convey information about commonly occurring
patterns that are present in images with positive tags and absent
in images with negative tags. The system is thus susceptible to strong
inter-class co-occurrences (e.g. train with rail), as well as systematic
part occlusions (e.g. feet).

\paragraph{\label{subsec:CRF-and-CRFLoss}CRF and CRFLoss}

A traditional approach to make labels match object boundaries is to
solve a CRF inference problem \cite{Lafferty2001ICML,Kraehenbuehl2011Nips}
over the image grid; where the pair-wise terms relate to the object boundaries.
CRF can be applied at three stages: (1) on the seeds (\texttt{crf-seed}),
(2) as a loss function during segmenter convnet training (\texttt{crf-loss})
\cite{kolesnikov2016seed}, and (3) as a post-processing at test time
(\texttt{crf-postproc}). We have experimented with multiple combinations
of those (see supplementary materials section \ref{sec:supp-crf}). 

Albeit some gains are observed, these are inconsistent. For example
$\mathtt{GAP}\text{-}\mathtt{HighRes}$ and $\mathtt{GAP}\text{-}\mathtt{ROI}$
provide near identical classification and seeding performance (see
table \ref{tab:GAP-architectures}), yet using the same CRF setup
provides $+13\ \mbox{mIoU}$ percent points in one, but only $+7\ \mbox{pp}$
on the other. In comparison our saliency approach will provide $+17\ \mbox{mIoU}$
and $+18\ \mbox{mIoU}$ for these two networks respectively (see below).

\subsection{\label{subsec:Saliency}Saliency}

\begin{figure*}
\begin{centering}
\subfloat[\label{fig:high-quality-saliency}High quality]{\begin{centering}
\includegraphics[width=0.15\textwidth,height=0.1\textwidth]{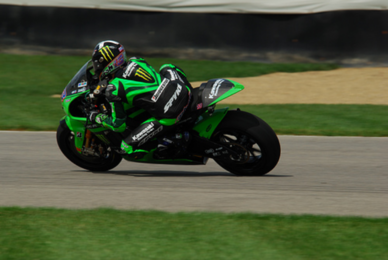}\,\includegraphics[width=0.15\textwidth,height=0.1\textwidth]{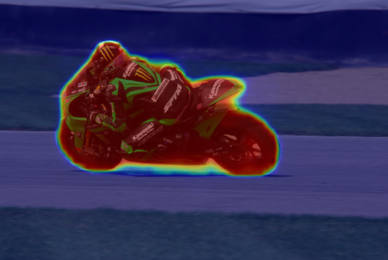}
\par\end{centering}
}~\subfloat[\label{fig:Medium-quality-saliency}Medium quality]{\begin{centering}
\includegraphics[width=0.15\textwidth,height=0.1\textwidth]{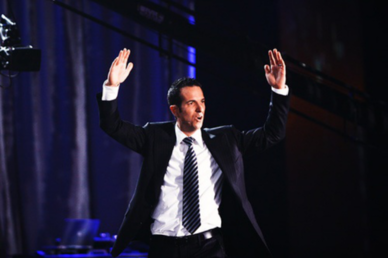}\,\includegraphics[width=0.15\textwidth,height=0.1\textwidth]{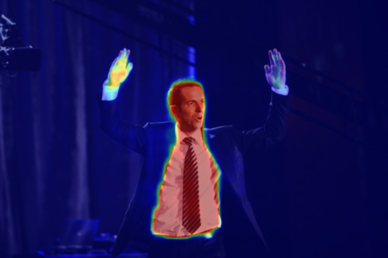}
\par\end{centering}
}~\subfloat[\label{fig:Low-quality-saliency}Low quality]{\begin{centering}
\includegraphics[width=0.15\textwidth,height=0.1\textwidth]{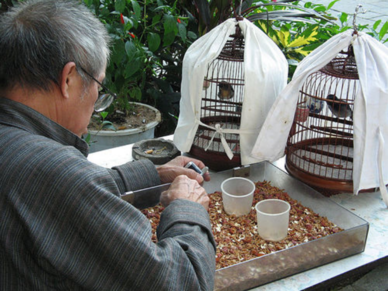}\,\includegraphics[width=0.15\textwidth,height=0.1\textwidth]{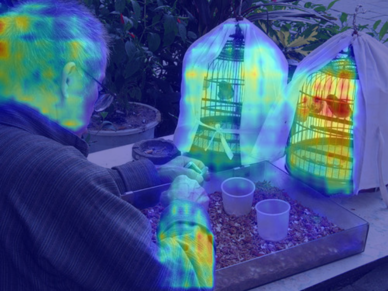}
\par\end{centering}
}\vspace{-1em}
\par\end{centering}
\caption{\label{fig:saliency-examples}Example of our saliency map results
on Pascal VOC 2012 data.}
\end{figure*}

\noindent
We propose to use object saliency to extract information about the
object extent. We work under the assumption that a large portion of
the dataset are intentional photographies, which is the case for most
datasets crawled from the web such as Pascal \cite{pascal-voc-2012}
and Coco \cite{Lin2014EccvCoco}. If the image contains a single label
``dog'', chances are that the image is about a dog, and that the
salient object of the image is a dog. We use a convnet based saliency
estimator (detailed in \S \ref{sec:Saliency-details}) which
adds the benefit of translation invariance. If two locally salient
dogs appear in the image, both will be labelled as foreground.

When using saliency to guide semantic labelling at least two difficulties
need to be handled. For one, saliency per-se does not segment object
instances. In the example figure \ref{fig:high-quality-saliency},
the person-bike is well segmented, but person and bike are not separated.
Yet, the ideal Guide labeller (fig. \ref{fig:high-level-architecture})
should give different labels to these two objects. The second difficulty,
clearly visible in the examples of figure \ref{fig:saliency-examples},
is that the salient object might not belong to a category of interest
(shirt instead of person in figure \ref{fig:Medium-quality-saliency})
or that the method fails to identify any salient region at all (figure
\ref{fig:Low-quality-saliency}). 

We measure the saliency quality when compared to the ground truth
foreground on Pascal VOC 2012 validation\emph{ }set. Albeit our convnet
saliency model is better than hand-crafted methods \cite{Jiang2013Cvpr,zhang2015MBD},
in the end only about $20\%$ of images have reasonably good ($\text{IoU}>0.6$)
foreground saliency quality. Yet, as we will see in \S \ref{sec:Experiments},
this bit of information is already helpful for the weakly supervised
learning task.

\begin{figure}
\begin{centering}
\begin{tabular}{cccc}
\begin{turn}{90}
\hspace*{-1em}%
\begin{tabular}{c}
Salient object\tabularnewline
{\small{}and boxes}\tabularnewline
\end{tabular}
\end{turn} & \includegraphics[height=0.2\columnwidth]{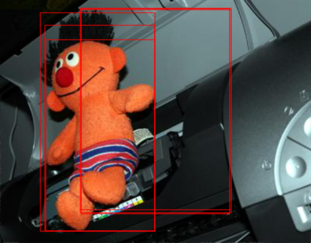}\hspace*{-0.5em} & \includegraphics[height=0.2\columnwidth]{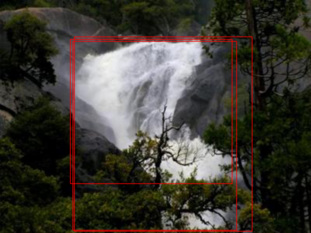}\hspace*{-0.5em} & \includegraphics[height=0.2\columnwidth]{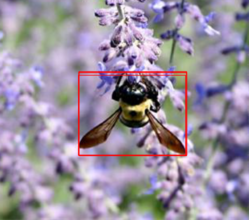}\hspace*{-0.5em}\tabularnewline
\begin{turn}{90}
\hspace*{-1em}%
\begin{tabular}{c}
Saliency \tabularnewline
model{\small{} result}\tabularnewline
\end{tabular}
\end{turn} & \includegraphics[height=0.2\columnwidth]{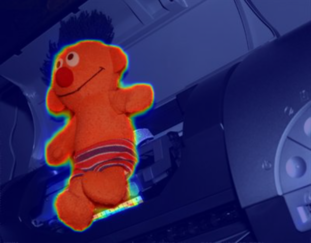}\hspace*{-0.5em} & \includegraphics[height=0.2\columnwidth]{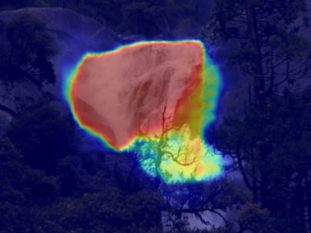}\hspace*{-0.5em} & \includegraphics[height=0.2\columnwidth]{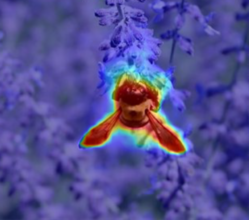}\hspace*{-0.5em}\tabularnewline
\end{tabular}
\par\end{centering}
\caption{\label{fig:MSRA-saliency}Example of saliency results on its training
data. We use MSRA box annotations to train a weakly supervised saliency
model. Note that the MSRA subset employed does not contain Pascal
categories.}
\end{figure}

 Crucially, our saliency system is trained on images containing diverse
 objects (hundreds of categories), the object categories are treated
 as ``unknown'', and to ensure clean experiments we handicap the
 system by removing any instance of Pascal categories in the object
 saliency training set. Our saliency model captures a general notion
 of plausible foreground objects and background areas (more details
 in section \ref{sec:Saliency-details}).
 
 On every Pascal training image, we obtain a class-agnostic foreground/background
 binary mask from our saliency model, and high precision/low recall
 class-specific image labels from the seeds model (section \ref{sec:goods-seeds}).
 We want to combine them in such a way that seed signals are well propagated
 throughout the foreground saliency mask. We consider two baselines
 strategies to generate guide labels using saliency but no seeds ($\mathcal{G}_{0}$
 and $\mathcal{G}_{1}$), and then discuss how we combine saliency
 with seeds ($\mathcal{G}_{2}$). 

\paragraph{$\mathcal{G}_{0}$ Random class assignment}

Given a saliency mask, we assign all foreground pixels to a class
randomly picked from the ground truth image labels. If a single ``dog''
label is present, then all foreground pixels are ``dog''. Two labels
are present (``dog, cat''), then all pixels are either dog or cat.

\paragraph{$\mathcal{G}_{1}$ Per-connected component classification}

Given a saliency mask, we split it in components, and assign a separate
label for each component. The per-component labels are given using
a full-image classifier trained using the image labels (classifier
details in \S \ref{par:G1-Classifier-details}). Given a connected
component mask $R_{i}^{fg}$ (with pixel values $1$: foreground,
$0$: background), we compute the classifier scores when feeding the
original image ($I$), and when feeding an image with background zeroed
($I\odot R_{i}^{fg}$). Region $R_{i}^{fg}$ will be labelled with
the ground truth class with the greatest positive score difference
before and after zeroing.

\paragraph{$\mathcal{G}_{2}$ Propagating seeds}

Here, instead of assigning the label per connected component $R_{i}^{fg}$
using a classifier, we instead use the seed labels. We also treat
the seeds as a set of connected components (seed $R_{j}^{s}$). Depending
on how the seeds and the foreground regions intersect, we decide the
label for each pixel in the guide labeller output. 

Our fusion strategy uses five simple ideas. 1) We treat the seeds
as reliable small size point predictors of each object instance, but
that might leak outside of the object. 2) We assume the saliency might
trigger on objects that are not part of the classes of interest. 3)
A foreground connected component $R_{i}^{fg}$ should take the label
of the seed touching it, 4) If two (or more) seeds touch the same
foreground component, then we want to propagate all the seed labels
inside it. 5) When in doubt, mark as ignore. The details for the corner
cases are provided in the supplementary material section \ref{sec:supp-alg}.

\begin{figure*}
\begin{centering}
\hspace*{\fill}\subfloat[Image]{\begin{centering}
\includegraphics[width=0.24\columnwidth]{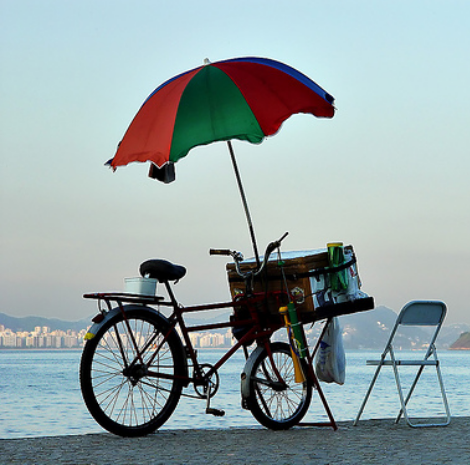}
\par\end{centering}
}\hspace*{\fill}\subfloat[Ground truth]{\begin{centering}
\includegraphics[width=0.24\columnwidth]{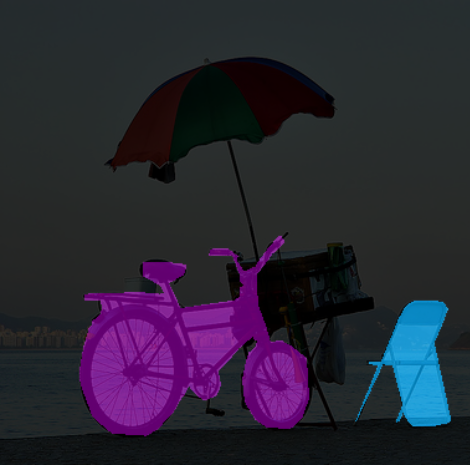}
\par\end{centering}
}\hspace*{\fill}\subfloat[Seed]{\begin{centering}
\includegraphics[width=0.24\columnwidth]{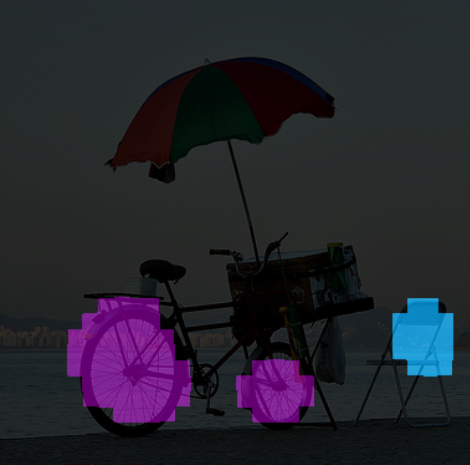}
\par\end{centering}
}\hspace*{\fill}\subfloat[Saliency]{\begin{centering}
\includegraphics[width=0.24\columnwidth]{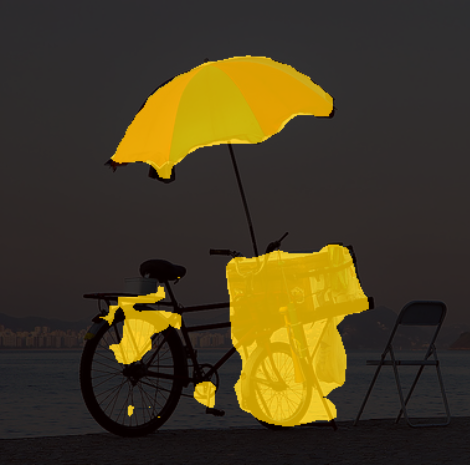}
\par\end{centering}
}\hspace*{\fill}\subfloat[$\mathcal{G}_{0}$]{\begin{centering}
\includegraphics[width=0.24\columnwidth]{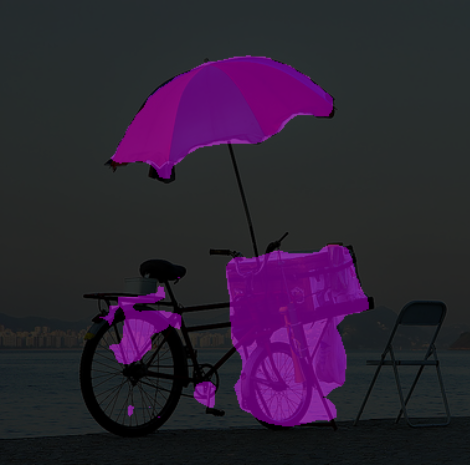}
\par\end{centering}
}\hspace*{\fill}\subfloat[$\mathcal{G}_{1}$]{\begin{centering}
\includegraphics[width=0.24\columnwidth]{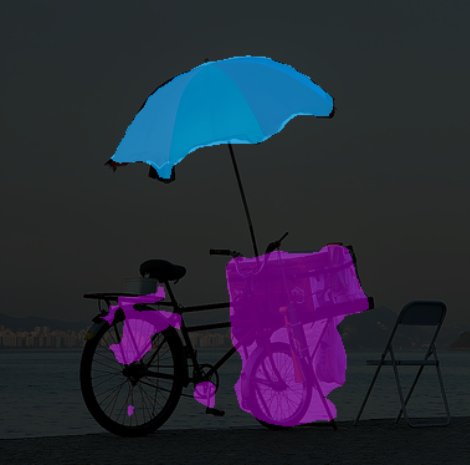}
\par\end{centering}
}\hspace*{\fill}\subfloat[$\mathcal{G}_{2}$]{\begin{centering}
\includegraphics[width=0.24\columnwidth]{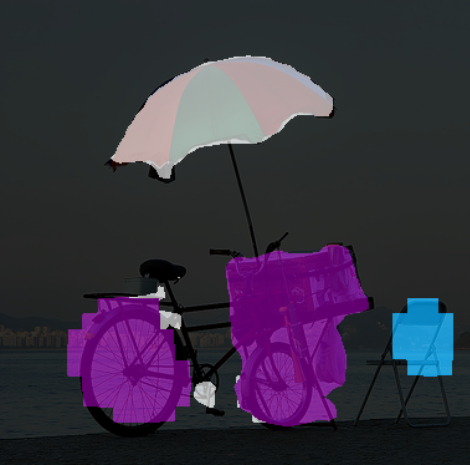}
\par\end{centering}
}\hspace*{\fill}
\par\end{centering}
\caption{\label{fig:combination-strategies-examples}Guide labelling strategies
example results. The image, its labels (``bicycle, chair''), seeds,
and saliency map are their input. White overlay indicates ``ignore''
pixel label.}
\end{figure*}

Figure \ref{fig:combination-strategies-examples} provides example
results of the different guide strategies. For additional qualitative
examples of seeds, saliency foreground, and generated labels, see
figure \ref{fig:qualitative-examples}. With our guide strategies
$\mathcal{G}_{0}$, $\mathcal{G}_{1}$, and $\mathcal{G}_{2}$ at
hand, we now proceed to empirically evaluate them in \S \ref{sec:Experiments}.

\section{\label{sec:Experiments}Experiments}

\noindent
\S \ref{subsec:Experimental-setup} and \ref{subsec:Implementation-details}
provide the details of the evaluation and our implementation. \S
\ref{subsec:Ingredients-study} compares our different guide strategies
amongst each other, and \S \ref{subsec:main-results} compares
with previous work on weakly supervised semantic labelling from image-level
labels.

\paragraph{\label{subsec:Experimental-setup}Evaluation}

We evaluate our image-level supervised semantic segmentation system
on the PASCAL VOC 2012 segmentation benchmark \cite{pascal-voc-2012}.
We report all the intermediate results on the \textit{val}\emph{ }set
($1\,449$ images) and only report the final system result on the
\textit{test} set ($1\,456$ images). Evaluation metric is the standard mean
intersection-over-union (mIoU).

\subsection{\label{subsec:Implementation-details}Implementation details}

\noindent
For training seeder and Segmenter networks, we use the
ImageNet \cite{imagenet_cvpr09} pretrained models for initialisation
and fine-tune on the Pascal VOC 2012 \emph{trainaug} set ($10\,582$
images), an extension of the original \textit{train} set ($1\,464$ images)
\cite{pascal-voc-2012,Hariharan2011Iccv}. This is the same procedure
used by previous work on fully \cite{Chen2016ArxivDeeplabv2} and
weakly supervised learning \cite{kolesnikov2016seed}.

\paragraph{Seeder}

Results in tables \ref{tab:pascal-results-ours} and \ref{tab:pascal-results-others}
are obtained using $\mathtt{GAP}\text{-}\mathtt{HighRes}$ (see \S\ref{sec:goods-seeds}),
trained for image classification on the Pascal trainaug set. The test
time foreground threshold $\tau$ is set to $0.2$, following the
previous literature \cite{zhou2015cnnlocalization,kolesnikov2016seed}.

\paragraph{$\mathcal{G}_{1}$ \label{par:G1-Classifier-details}Classifier}

The guide labeller strategy $\mathcal{G}_{1}$ uses an image classifier
trained on Pascal \textit{trainaug} set. We use the VGG-16 architecture \cite{Simonyan2015Iclr}
with a multi-label loss. 

\paragraph{\label{sec:Saliency-details}Saliency}

Following \cite{Zhao2015Cvpr,Li2016Tip,Li2016Cvpr} we re-purpose
a semantic labelling network for the task of class-agnostic saliency.
 We train the DeepLab-v2 ResNet \cite{Chen2016ArxivDeeplabv2} over
  a subset of MSRA \cite{liu2011learning}, a saliency dataset with
  \emph{class agnostic} bounding box annotations. We constrain the training
  only to samples of \emph{non-Pascal} categories. Thus, the saliency
  model does not leverage class specific features when Pascal images
  are fed. Out of $25k$ MSRA images, $11\,041$ remain after filtering. 
  
MRSA provides bounding boxes (from multiple annotators) of the main
salient element of each image. To train the saliency model to output
pixel-wise masks, we follow \cite{Khoreva2016Arxiv}. We generate
segments from the MSRA boxes by applying grabcut over the average
box annotation, and use these as supervision for the DeepLab model.
The model is trained as a binary semantic labeller for foreground
and background regions. The trained model generates masks like the
ones shown in figure \ref{fig:MSRA-saliency}. Although having been
trained with images with single salient objects, due to its convolutional
nature the network can predict multiple salient regions in the Pascal
images (as shown in figure \ref{fig:qualitative-examples}).

At test time, the saliency model generates a heatmap of foreground
probabilities. We threshold at $50\%$ of the maximal foreground
probability to generate the mask.

\paragraph{Segmenter}

For comparison with previous work we use the DeepLabv1-LargeFOV \cite{Chen2016ArxivDeeplabv2}
architecture as our segmenter convnet. The network is trained on Pascal
\textit{trainaug} set with $10\,582$ images, using the output of the guide
labeller (\S\ref{fig:high-level-architecture}), which uses only
the image itself and presence-absence tags of the $20$ Pascal categories
as supervision. The network is trained for $8k$ iterations. \\
Following the standard DeepLab procedure, at test time we up-sample
the output to the original image resolution and apply the dense CRF
inference \cite{Kraehenbuehl2011Nips}. Unless stated otherwise, we
use the CRF parameters used for DeepLabv1-LargeFOV \cite{Chen2016ArxivDeeplabv2}.

Additional training details and hyper-parameters are given in the
supplementary materials section \ref{sec:supp-convnet-training}.

\subsection{\label{subsec:Ingredients-study}Ingredients study}

 \noindent Table \ref{tab:pascal-results-ours} compares different
 guide strategies $\mathcal{G}_{0}$, $\mathcal{G}_{1}$, $\mathcal{G}_{2}$,
 and oracle versions of $\mathcal{G}_{2}$. The first row shows the
 result of training our segmenter using the seeds directly as guide
 labels. This leads to poor quality ($38.7\ \mbox{mIoU}$). The ``Supervision''
column shows recall and precision for foreground and background of
the guide labels themselves (training data for the segmenter). We
can see that the seeds alone have low recall for the foreground ($37\%$).
In comparison, using saliency only, $\mathcal{G}_{0}$ reaches significantly
better results, due to the higher foreground recall ($52\%$), at
a comparable precision.

Adding a classifier on top of the saliency ($\mathcal{G}_{0}\rightarrow\mathcal{G}_{1}$)
provides only a negligible improvement ($45.8\rightarrow46.2$). This
can be attributed the fact that many Pascal images contain only a
single foreground class, and that the classifier might have difficulties
recognizing the masked objects. Interestingly, when using a similar
classifier to generate seeds instead of scoring the image ($\mathcal{G}_{1}\rightarrow\mathcal{G}_{2}$)
we gain $5\ \text{pp}$ (percent points, $46.2\rightarrow51.2$).
This shows that the details of how a classifier is used can make a
large difference. 

Table \ref{tab:pascal-results-ours} also reports a saliency oracle
case on top of $\mathcal{G}_{2}$. If we use the ground truth annotation
 to generate an ideal saliency mask, we see a significant improvement
 over $\mathcal{G}_{2}$ ($51.2\rightarrow56.9$). Thus, the quality
 of saliency is an important ingredient, and there is room for further
 gains.

\begin{table}
\begin{centering}
\setlength{\tabcolsep}{1pt} 
\par\end{centering}
\begin{centering}
\par\end{centering}
\begin{centering}
\hspace*{-1em}%
\begin{tabular}{ccc|cccc|c}
\multirow{2}{*}{Method} & \multirow{2}{*}{Seeds} & Sali- & \multicolumn{4}{c|}{Supervision} & \multicolumn{1}{c}{val. set}\tabularnewline
 &  & ency & \multicolumn{2}{c}{Fg {\small{}P/R}} & \multicolumn{2}{c|}{Bg {\small{}P/R}} & mIoU\tabularnewline
\hline 
\hline 
Seeds only & $\checkmark$ & \textbf{\scriptsize{}\XSolidBrush{}} & 69 & 37 & 81 & 95 & 38.7\tabularnewline
$\mathcal{G}_{0}$ & \textbf{\scriptsize{}\XSolidBrush{}} & $\checkmark$ & 65 & 52 & 65 & 52 & 45.8\tabularnewline
$\mathcal{G}_{1}$  & \textbf{\scriptsize{}\XSolidBrush{}} & $\checkmark$ & 75 & 51 & 75 & 51 & 46.2\tabularnewline
$\mathcal{G}_{2}$ & $\checkmark$ & $\checkmark$ & 73 & 59 & 87 & 95 & 51.2\tabularnewline
\hline 
\multirow{1}{*}{\hspace*{-1em}Saliency oracle\hspace*{-2em}} & $\checkmark$ & $\checkmark$ & 89 & 91 & 100 & 99 & 56.9\tabularnewline
\end{tabular}
\par\end{centering}
\begin{centering}
\par\end{centering}
\caption{\label{tab:pascal-results-ours}Comparison of different guide labeller
variants. Pascal VOC 2012 validation set results, without CRF post-processing.
Fg/Bg P/R: are foreground/background precision and recall of the guide
labels. Discussion in \S\ref{subsec:Ingredients-study}.}
\end{table}

\subsection{\label{subsec:main-results}Results}

 \noindent Table \ref{tab:pascal-results-others} compares our results
 with previous related work. We group results by methods that only
 use ImageNet pre-training and image-level labels (I, P, E; see legend
 table \ref{tab:pascal-results-others}), and methods that use additional
 data or user-inputs. Here our $\mathcal{G}_{0}$ and $\mathcal{G}_{2}$
 results include a CRF post-processing (\texttt{crf-postproc}). We
 also experimented with \texttt{crf-loss }but did not find a parameter
set that provided improved results.

We see that the guide strategies $\mathcal{G}_{0}$, which uses saliency
and random ground-truth label, reaches competitive performance compared
to methods using $\text{I}$+P only. This shows that saliency by itself
is already a strong cue. Our guide strategy $\mathcal{G}_{2}$ (which
uses seeds and saliency) obtains the best reported results on this
task\footnote{\cite{Qi2016Eccv} also reports $54.3$ validation set results, however
we do not consider these results comparable since they use the MCG
scores \cite{PontTuset2015ArxivMcg}, which are trained on the ground
truth Pascal segments.}. We even improve over other methods using saliency (\texttt{STC})
or using additional human annotations (\texttt{MicroAnno},\texttt{
CheckMask}). Compared to a fully supervised DeepLabv1 model, our results
reach $80\%$ of the fully supervised quality. 

\begin{table}
\begin{centering}
\begingroup
\setlength{\tabcolsep}{0.15em} 
\begin{tabular}{cccccc}
 &  &  & val. set & \multicolumn{2}{c}{test set}\tabularnewline
 & Method & Data & mIoU & mIoU & $\text{FS}\%$\tabularnewline
\hline 
\hline 
\multirow{9}{*}{\begin{turn}{90}
{\small{}Image labels only}
\end{turn}} & \texttt{MIL-FCN}\,\cite{Pathak2015Iclrw} & $\text{I}$+P & 25.0 & 25.6 & 36.5\tabularnewline
 & \texttt{CCNN}\,\cite{Pathak2015Iccv} & $\text{I}$+P & 35.3 & 35.6 & 50.6\tabularnewline
 & \texttt{WSSL}\,\cite{Papandreou2015Iccv} & $\text{I}$+P & 38.2 & 39.6 & 56.3\tabularnewline
 & \texttt{MIL+Seg}\,\cite{Pinheiro2015Cvpr} & $\text{I}$+$\text{E}_{760k}$ & 42.0 & 40.6 & 57.8\tabularnewline
 & \texttt{DCSM}\,\cite{Shimoda2016Eccv} & $\text{I}$+P & 44.1 & 45.1 & 64.2\tabularnewline
 & \texttt{CheckMask}\,\cite{Saleh2016Eccv} & $\text{I}$+P & 46.6 & - & -\tabularnewline
 & \texttt{SEC}\,\cite{kolesnikov2016seed} & $\text{I}$+P & 50.7 & 51.7 & 73.5\tabularnewline
 & \texttt{AF-ss}\,\cite{Qi2016Eccv} & $\text{I}$+P & 51.6 & - & -\tabularnewline
\cline{2-6} 
 & Seeds only & $\text{I}$+P & 39.8 & - & -\tabularnewline
\hline 
\multirow{6}{*}{\begin{turn}{90}
{\small{}More information}
\end{turn}} & \texttt{CCNN}\,\cite{Pathak2015Iccv} & $\text{I}$+P+Z & - & 45.1 & 64.2\tabularnewline
 & \texttt{STC}\,\cite{Wei2015ArXiv} & $\text{I}$+P+$\text{S}$+$\text{E}_{40k}$ & 49.8 & 51.2 & 72.8\tabularnewline
 & \texttt{CheckMask}\,\cite{Saleh2016Eccv} & $\text{I}$+P+$\mu$ & 51.5 & - & -\tabularnewline
 & \texttt{MicroAnno}\,\cite{Kolesnikov2016Bmvc} & $\text{I}$+P+$\mu$ & 51.9 & 53.2 & 75.7\tabularnewline
\cline{2-6} 
 & $\mathcal{G}_{0}$ & $\text{I}$+P+$\text{S}$ & 48.8 & - & -\tabularnewline
 & $\mathcal{G}_{2}$ & $\text{I}$+P+$\text{S}$ & \textbf{55.7} & \textbf{56.7} & \textbf{80.6}\tabularnewline
\hline 
 & DeepLabv1 & $\text{I}$+$\text{P}_{full}$ & 67.6 & 70.3 & 100\tabularnewline
\end{tabular}\endgroup
\par\end{centering}
\caption{\label{tab:pascal-results-others}Comparison of state of the art methods,
on Pascal VOC 2012 val.\emph{ }and test set. $\text{FS}\%$: fully
supervised percent. Ingredients: $\text{I}$: ImageNet classification
pre-training, P: Pascal image level tags, $\text{P}_{full}$: fully
supervised case (pixel wise labels), $\text{E}_{n}$: $n$ extra images
with image level tags, S: saliency, Z: per-class size prior, $\mu$:
human-in-the-loop micro-annotations.}
\end{table}

\begin{figure}
\begin{centering}
\begin{tabular}{ccc}
\begin{turn}{90}
Input image
\end{turn} & \includegraphics[width=0.35\columnwidth]{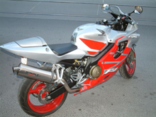} & \includegraphics[width=0.47\columnwidth]{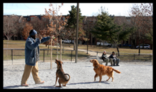}\tabularnewline
\begin{turn}{90}
Seeds\hspace*{-3em}
\end{turn} & \includegraphics[width=0.35\columnwidth]{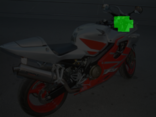} & \includegraphics[width=0.47\columnwidth]{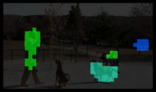}\tabularnewline
\begin{turn}{90}
Saliency\hspace*{-3em}
\end{turn} & \includegraphics[width=0.35\columnwidth]{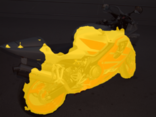} & \includegraphics[width=0.47\columnwidth]{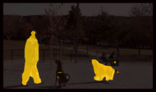}\tabularnewline
\begin{turn}{90}
\begin{tabular}{c}
$\mathcal{G}_{2}$\tabularnewline
{\small{}(Seeds+Saliency)}\tabularnewline
\end{tabular}
\end{turn} & \includegraphics[width=0.35\columnwidth]{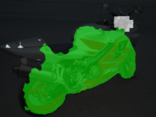} & \includegraphics[width=0.47\columnwidth]{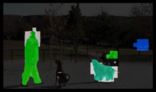}\tabularnewline
\begin{turn}{90}
\begin{tabular}{c}
Segmenter\tabularnewline
output\tabularnewline
\end{tabular}
\end{turn} & \includegraphics[width=0.35\columnwidth]{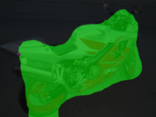} & \includegraphics[width=0.47\columnwidth]{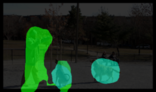}\tabularnewline
\begin{turn}{90}
+CRF\hspace*{-3em}
\end{turn} & \includegraphics[width=0.35\columnwidth]{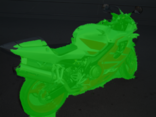} & \includegraphics[width=0.47\columnwidth]{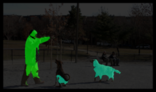}\tabularnewline
\begin{turn}{90}
Ground truth
\end{turn} & \includegraphics[width=0.35\columnwidth]{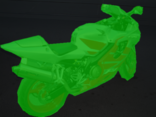} & \includegraphics[width=0.47\columnwidth]{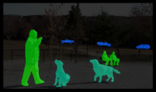}\tabularnewline
\end{tabular}
\par\end{centering}
\caption{\label{fig:qualitative-examples}Qualitative examples of the different
stages of our system. Additional examples in the supplementary
materials figures \ref{fig:qualitative-examples-1} and \ref{fig:qualitative-examples-2}.}
\end{figure}

\section{\label{sec:Conclusion}Conclusion}

\noindent
We have addressed the problem of training a semantic segmentation
convnet from image labels. Image labels alone can provide high quality
seeds, or discriminative object regions, but learning the full object
extents is a hard problem. We have shown that saliency is a viable
option for feeding the object extent information.

The proposed Guided Segmentation architecture (\S\ref{sec:architecture}),
where the ``guide labeller'' combines cues from the seeds and saliency,
can successfully train a segmentation convnet to achieve the state
of the art performance. Our weakly supervised results reach $80\%$
of the fully supervised case.

We expect that a deeper understanding of the seeder methods and improvements
on the saliency model can lead to further improvements. 

\subsection*{Acknowledgements}

\noindent
This research was supported by the German Research Foundation (DFG CRC
1223).


\bibliographystyle{ieee}
\bibliography{arxiv}

\newpage
\appendix


\part*{Supplementary Materials}

\section{\label{sec:supp-content}Content}

This document contains the following additional details:
\begin{itemize}
\item GAP

\begin{itemize}
\item Architecture details for $\mathtt{GAP}\text{-}\mathtt{LowRes}$, $\mathtt{GAP}\text{-}\mathtt{HighRes}$,
$\mathtt{GAP}\text{-}\mathtt{DeepLab}$, $\mathtt{GAP}\text{-}\mathtt{ROI}$.
\item Training details.
\item Qualitative results.
\end{itemize}
\item CRF experiments and CRF parameters used.
\item More qualitative results of our saliency model.
\item Details of $\mathcal{G}_{2}$ guide labeller rules, and more qualitative
examples of $\mathcal{G}_{0}$, $\mathcal{G}_{1}$, and $\mathcal{G}_{2}$
strategies.
\item Convnet training details for Seeder, Classifier, and Segmenter networks.
\item Additional qualitative examples like figure \ref{fig:qualitative-examples} in the main paper.
\end{itemize}

\section{\label{sec:supp-gap}GAP}

\subsection{\label{sec:supp-gap-net}Network details}

See table \ref{tab:GAP-networks} for the details of the GAP networks
used. For $\mathtt{GAP}\text{-}\mathtt{ROI}$, we insert the GAP layer
after the final linear layer, instead of after the penultimate layer
as suggested by \cite{zhou2015cnnlocalization}. We note that the
resulting functions are identical (and hence the forward and backward
passes): GAP is a linear sum over the spatial dimensions, and the
final layer performs a linear combination over the channel dimensions,
so they can be swapped without changing the function. Also in practice,
we find that there is only negligible difference in performance between
the variants.

\subsection{\label{sec:supp-gap-train}Training GAP}

All GAP network variants are trained with stochastic gradient descent
(SGD) with minibatch size $15$, momentum $0.9$, weight decay $5\times10^{-4}$,
and base learning rate $0.001$, decreased by the factor of $10$
at every $2\,000$ iterations. The training stops at $8\,000$ iterations.

\subsection{\label{sec:supp-gap-qual}Qualitative examples}

\begin{figure*}
\begin{centering}
\begin{tabular}{cccccc}

Image & $\mathtt{GAP}\text{-}\mathtt{LowRes}$ &  $\mathtt{GAP}\text{-}\mathtt{HighRes}$ &  $\mathtt{GAP}\text{-}\mathtt{ROI}$ &  $\mathtt{GAP}\text{-}\mathtt{DeepLab}$ & Ground truth\tabularnewline
\includegraphics[width=0.3\columnwidth]{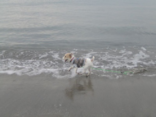} & \includegraphics[width=0.3\columnwidth]{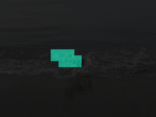} & \includegraphics[width=0.3\columnwidth]{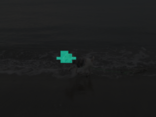} & \includegraphics[width=0.3\columnwidth]{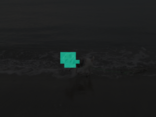} & \includegraphics[width=0.3\columnwidth]{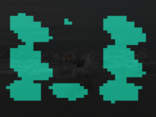} & \includegraphics[width=0.3\columnwidth]{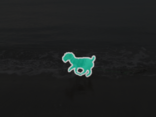}\tabularnewline
\includegraphics[width=0.3\columnwidth]{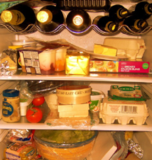} & \includegraphics[width=0.3\columnwidth]{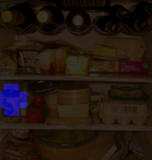} & \includegraphics[width=0.3\columnwidth]{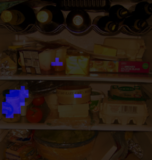} & \includegraphics[width=0.3\columnwidth]{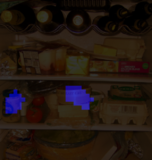} & \includegraphics[width=0.3\columnwidth]{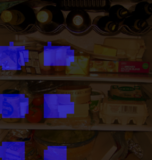} & \includegraphics[width=0.3\columnwidth]{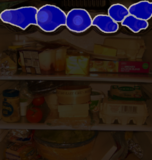}\tabularnewline
\includegraphics[width=0.3\columnwidth]{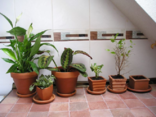} & \includegraphics[width=0.3\columnwidth]{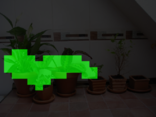} & \includegraphics[width=0.3\columnwidth]{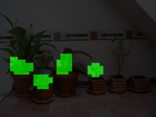} & \includegraphics[width=0.3\columnwidth]{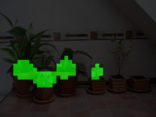} & \includegraphics[width=0.3\columnwidth]{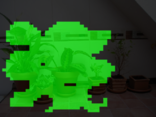} & \includegraphics[width=0.3\columnwidth]{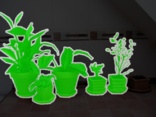}\tabularnewline
\includegraphics[width=0.3\columnwidth]{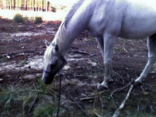} & \includegraphics[width=0.3\columnwidth]{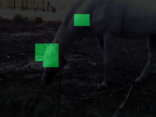} & \includegraphics[width=0.3\columnwidth]{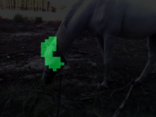} & \includegraphics[width=0.3\columnwidth]{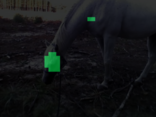} & \includegraphics[width=0.3\columnwidth]{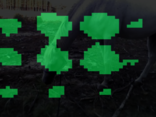} & \includegraphics[width=0.3\columnwidth]{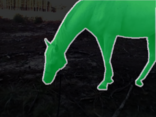}\tabularnewline
\includegraphics[width=0.3\columnwidth]{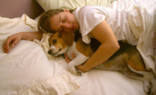} & \includegraphics[width=0.3\columnwidth]{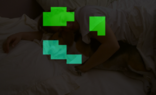} & \includegraphics[width=0.3\columnwidth]{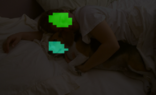} & \includegraphics[width=0.3\columnwidth]{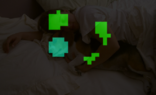} & \includegraphics[width=0.3\columnwidth]{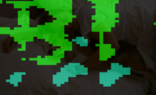} & \includegraphics[width=0.3\columnwidth]{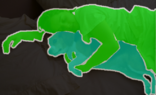}\tabularnewline
\includegraphics[width=0.3\columnwidth]{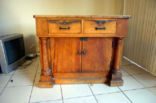} & \includegraphics[width=0.3\columnwidth]{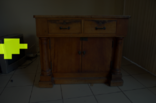} & \includegraphics[width=0.3\columnwidth]{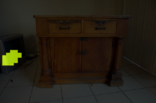} & \includegraphics[width=0.3\columnwidth]{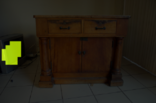} & \includegraphics[width=0.3\columnwidth]{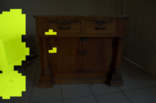} & \includegraphics[width=0.3\columnwidth]{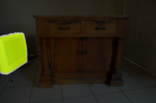}\tabularnewline
\includegraphics[width=0.3\columnwidth]{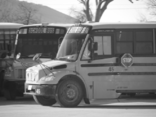} & \includegraphics[width=0.3\columnwidth]{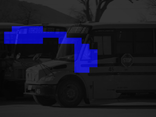} & \includegraphics[width=0.3\columnwidth]{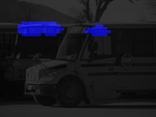} & \includegraphics[width=0.3\columnwidth]{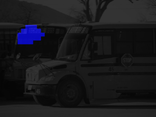} & \includegraphics[width=0.3\columnwidth]{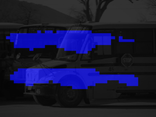} & \includegraphics[width=0.3\columnwidth]{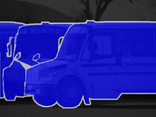}\tabularnewline
\includegraphics[width=0.3\columnwidth]{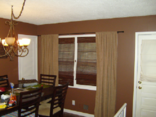} & \includegraphics[width=0.3\columnwidth]{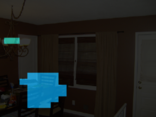} & \includegraphics[width=0.3\columnwidth]{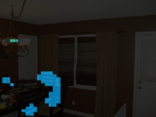} & \includegraphics[width=0.3\columnwidth]{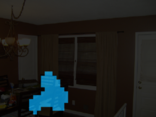} & \includegraphics[width=0.3\columnwidth]{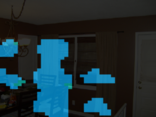} & \includegraphics[width=0.3\columnwidth]{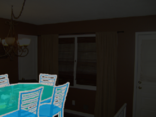}\tabularnewline
 &  &  &  &  & \tabularnewline
\end{tabular}
\par\end{centering}
\caption{\label{fig:GAP-qualitative}Qualitative examples of GAP output for
$\mathtt{GAP}\text{-}\mathtt{LowRes}$, $\mathtt{GAP}\text{-}\mathtt{HighRes}$,
$\mathtt{GAP}\text{-}\mathtt{DeepLab}$, and $\mathtt{GAP}\text{-}\mathtt{ROI}$.
Note that all of them, except for $\mathtt{GAP}\text{-}\mathtt{DeepLab}$,
are qualitatively similar. For $\mathtt{GAP}\text{-}\mathtt{DeepLab}$,
we observe repeating patterns of certain stride. Examples are chosen
at random.}
\end{figure*}

See figure \ref{fig:GAP-qualitative} for qualitative examples. We
observe that $\mathtt{GAP}\text{-}\mathtt{LowRes}$, $\mathtt{GAP}\text{-}\mathtt{HighRes}$,
and $\mathtt{GAP}\text{-}\mathtt{ROI}$ show qualitatively similar
results, while $\mathtt{GAP}\text{-}\mathtt{DeepLab}$ has significantly
low quality with repeating patterns in the output. The output suggests
that the learned filters have repeating patterns modulo $\approx12$
output pixels, which is the width of the dilated filters in DeepLab-LargeFOV
\cite{Chen2016ArxivDeeplabv2} on $\mathtt{conv5}$ features.

\section{\label{sec:supp-crf}CRF}

See table \ref{tab:CRF-loss-summary} for the segmenter performance
after applying different combinations of CRF units, \texttt{crf-seed},
\texttt{crf-loss}, and \texttt{crf-postproc}. Combination of \texttt{crf-loss}
and \texttt{crf-postproc} on the $\mathtt{GAP}\text{-}\mathtt{HighRes}$
seed gives 50.4 mIoU, giving $12.9$ mIoU boost over the vanilla seed.
However, we do not see such a gain when either the CRF parameters
or the seed type is changed. When CRF parameters are changed from
$v_{1}$ to $v_{2}$, both of which are reasonable choices (see \S\ref{sec:crf-parameters}),
we lose $5.2$ mIoU. When the seed type is changed from $\mathtt{GAP}\text{-}\mathtt{HighRes}$
to $\mathtt{GAP}\text{-}\mathtt{ROI}$, we lose $5.4$ mIoU. The $12.9$
mIoU boost thus seems fragile.

Our saliency-based model, on the other hand, gives a consistent $\geq4$
mIoU gain over the best CRF combination, regardless of the seed type
used, showing superiority over CRF both in terms of performance and
stability. It is possible to combine \texttt{crf-loss} and saliency,
but our preliminary experiments show that it hurts the performance
of the saliency-only case. Thus, \texttt{crf-loss} is excluded from
our final model.

See table \ref{tab:CRF-loss} for all the combinations considered
in our experiments. 

\begin{table*}
\caption{\label{tab:CRF-loss-summary}Results of the CRF variants on Pascal
2012 validation. $v_{1}$, $v_{2}$: CRF parameters from \cite{kolesnikov2016seed}
and \cite{Chen2016ArxivDeeplabv2} respectively.}

\begin{centering}
\begin{tabular}{cccccc}
 & \multicolumn{3}{c}{} & \multicolumn{2}{c}{}\tabularnewline
\hline 
Seed method & \multicolumn{3}{c}{\texttt{crf}} & \multicolumn{2}{c}{val. set}\tabularnewline
 & \texttt{-seed} & \texttt{-loss} & \texttt{-postproc} & mIoU & $\Delta\text{mIoU}$\tabularnewline
\hline 
$\mathtt{GAP}\text{-}\mathtt{HighRes}$ & \textbf{\scriptsize{}\XSolidBrush{}} & \textbf{\scriptsize{}\XSolidBrush{}} & \textbf{\scriptsize{}\XSolidBrush{}} & 37.5 & $-12.9$\tabularnewline
 & \textbf{\scriptsize{}\XSolidBrush{}} & $\checkmark_{v1}$ & $\checkmark_{v1}$ & 50.4 & 0\tabularnewline
 & \textbf{\scriptsize{}\XSolidBrush{}} & $\checkmark_{v2}$ & $\checkmark_{v2}$ & 45.2 & $-5.2$\tabularnewline
 & \multicolumn{3}{c}{Saliency: $\mathcal{G}_{2}$} & $55.2$ & $+4.8$\tabularnewline
\hline 
$\mathtt{GAP}\text{-}\mathtt{ROI}$ & \textbf{\scriptsize{}\XSolidBrush{}} & \textbf{\scriptsize{}\XSolidBrush{}} & \textbf{\scriptsize{}\XSolidBrush{}} & 37.6 & $-12.8$\tabularnewline
 & \textbf{\scriptsize{}\XSolidBrush{}} & $\checkmark_{v1}$ & $\checkmark_{v1}$ & 45.0 & $-5.4$\tabularnewline
 & \multicolumn{3}{c}{Saliency: $\mathcal{G}_{2}$} & $54.6$ & $+4.2$\tabularnewline
\hline 
 &  &  &  &  & \tabularnewline
\end{tabular}
\par\end{centering}

\end{table*}

\begin{table*}
\caption{\label{tab:CRF-loss}Extension of table \ref{tab:CRF-loss-summary}
showing all the combinations considered.}

\begin{centering}
\begin{tabular}{cccccc}
 & \multicolumn{3}{c}{} & \multicolumn{2}{c}{}\tabularnewline
\hline 
Seed method & \multicolumn{3}{c}{\texttt{crf}} & \multicolumn{2}{c}{val. set}\tabularnewline
 & \texttt{-seed} & \texttt{-loss} & \texttt{-postproc} & mIoU & $\Delta\text{mIoU}$\tabularnewline
\hline 
$\mathtt{GAP}\text{-}\mathtt{HighRes}$ & \textbf{\scriptsize{}\XSolidBrush{}} & \textbf{\scriptsize{}\XSolidBrush{}} & \textbf{\scriptsize{}\XSolidBrush{}} & 37.5 & $-12.9$\tabularnewline
 & \textbf{\scriptsize{}\XSolidBrush{}} & $\checkmark_{v1}$ & $\checkmark_{v1}$ & 50.4 & 0\tabularnewline
 & \textbf{\scriptsize{}\XSolidBrush{}} & $\checkmark_{v1}$ & \textbf{\scriptsize{}\XSolidBrush{}} & 46.4 & $-4.0$\tabularnewline
 & \textbf{\scriptsize{}\XSolidBrush{}} & \textbf{\scriptsize{}\XSolidBrush{}} & $\checkmark_{v1}$ & 45.5 & $-4.9$\tabularnewline
 & \textbf{\scriptsize{}\XSolidBrush{}} & $\checkmark_{v2}$ & $\checkmark_{v2}$ & 45.2 & $-5.2$\tabularnewline
 & $\checkmark_{v1}$ & \textbf{\scriptsize{}\XSolidBrush{}} & \textbf{\scriptsize{}\XSolidBrush{}} & 33.0 & $-17.4$\tabularnewline
 & \multicolumn{3}{c}{Saliency: $\mathcal{G}_{2}$} & $55.2$ & $+4.8$\tabularnewline
\hline 
$\mathtt{GAP}\text{-}\mathtt{ROI}$ & \textbf{\scriptsize{}\XSolidBrush{}} & \textbf{\scriptsize{}\XSolidBrush{}} & \textbf{\scriptsize{}\XSolidBrush{}} & 37.6 & $-12.8$\tabularnewline
 & \textbf{\scriptsize{}\XSolidBrush{}} & $\checkmark_{v1}$ & $\checkmark_{v1}$ & 45.0 & $-5.4$\tabularnewline
 & \textbf{\scriptsize{}\XSolidBrush{}} & $\checkmark_{v2}$ & $\checkmark_{v2}$ & 44.2 & $-6.2$\tabularnewline
 & \multicolumn{3}{c}{Saliency: $\mathcal{G}_{2}$} & $54.6$ & $+4.2$\tabularnewline
\hline 
 &  &  &  &  & \tabularnewline
\end{tabular}
\par\end{centering}

\end{table*}

\subsection{\label{sec:supp-crf-param}CRF parameters\label{sec:crf-parameters}}

Throughout the paper, we use the CRF parameters from the DeepLab-LargeFOV
model \cite{Chen2016ArxivDeeplabv2}, unless stated otherwise. The
parameters are given by $w^{(1)}=4$, $\theta_{\alpha}=121$, and
$\theta_{\beta}=5$ for the appearance kernel, and $w^{(2)}=3$ and
$\theta_{\gamma}=3$ for the smoothness kernel, following the notation
of equation 3 in \cite{Kraehenbuehl2011Nips}. For compatibility,
we always use the Potts model: $\mu(x_{i},x_{j})=1_{x_{i}=x_{j}}$.

For some experiments, we also use parameters from \cite{kolesnikov2016seed},
which uses $w^{(1)}=10$, $\theta_{\alpha}=80$, and $\theta_{\beta}=13$
for the appearance kernel, and $w^{(2)}=3$ and $\theta_{\gamma}=3$
for the smoothness kernel.

\section{\label{sec:supp-saliency}Saliency}

\begin{figure*}
\begin{centering}
\begin{tabular}{ccc|ccc}
Salient objects with boxes & Saliency model result &  &  & Salient objects with boxes & Saliency model result\tabularnewline
 &  &  &  &  & \tabularnewline
 &  &  &  &  & \tabularnewline
\includegraphics[width=0.4\columnwidth]{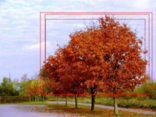} & \includegraphics[width=0.4\columnwidth]{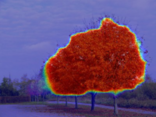} &  &  & \includegraphics[width=0.4\columnwidth]{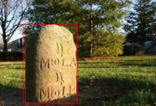} & \includegraphics[width=0.4\columnwidth]{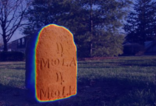}\tabularnewline
\includegraphics[width=0.4\columnwidth]{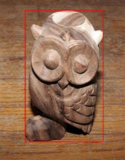} & \includegraphics[width=0.4\columnwidth]{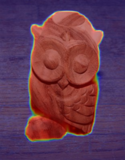} &  &  & \includegraphics[width=0.4\columnwidth]{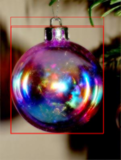} & \includegraphics[width=0.4\columnwidth]{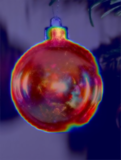}\tabularnewline
\includegraphics[width=0.4\columnwidth]{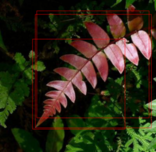} & \includegraphics[width=0.4\columnwidth]{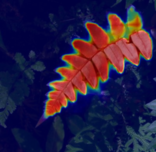} &  &  & \includegraphics[width=0.4\columnwidth]{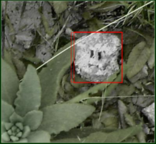} & \includegraphics[width=0.4\columnwidth]{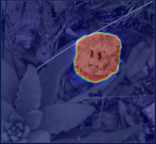}\tabularnewline
\includegraphics[width=0.4\columnwidth]{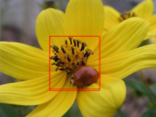} & \includegraphics[width=0.4\columnwidth]{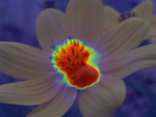} &  &  & \includegraphics[width=0.4\columnwidth]{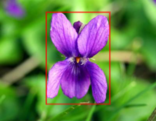} & \includegraphics[width=0.4\columnwidth]{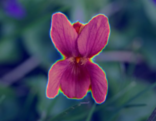}\tabularnewline
\includegraphics[width=0.4\columnwidth]{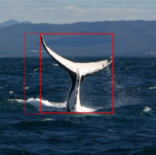} & \includegraphics[width=0.4\columnwidth]{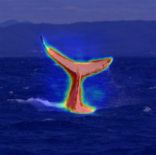} &  &  & \includegraphics[width=0.4\columnwidth]{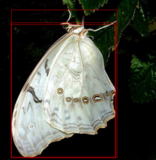} & \includegraphics[width=0.4\columnwidth]{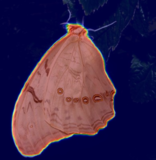}\tabularnewline
\includegraphics[width=0.4\columnwidth]{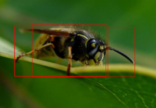} & \includegraphics[width=0.4\columnwidth]{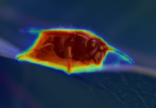} &  &  & \includegraphics[width=0.4\columnwidth]{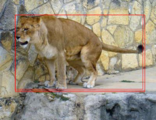} & \includegraphics[width=0.4\columnwidth]{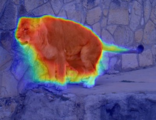}\tabularnewline
 &  & \multicolumn{1}{c}{} &  &  & \tabularnewline
\end{tabular}
\par\end{centering}
\caption{\label{fig:MSRA-saliency-1}Extension of figure \ref{fig:MSRA-saliency} in the main paper.
Examples of saliency results on its training data. We use MSRA box
annotations to train a weakly supervised saliency model. Note that
the MSRA subset employed is not biased towards the Pascal categories.
Examples are chosen at random.}
\end{figure*}

See figure \ref{fig:MSRA-saliency-1} for more examples of the MSRA
training samples for our weakly supervised saliency model. Samples
corresponding to Pascal categories are excluded from the training.

\begin{figure*}
\begin{centering}
\begin{tabular}{ccc|ccc}
\includegraphics[width=0.4\columnwidth]{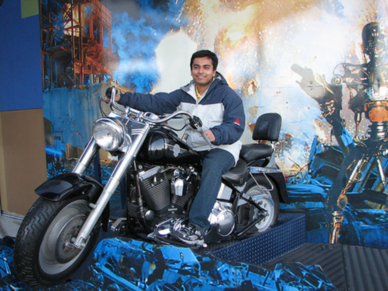} & \includegraphics[width=0.4\columnwidth]{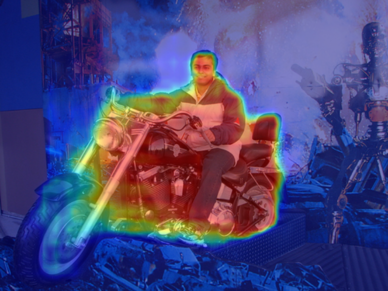} &  &  & \includegraphics[width=0.4\columnwidth]{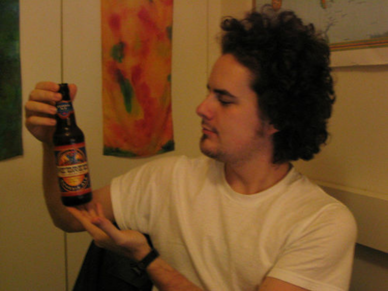} & \includegraphics[width=0.4\columnwidth]{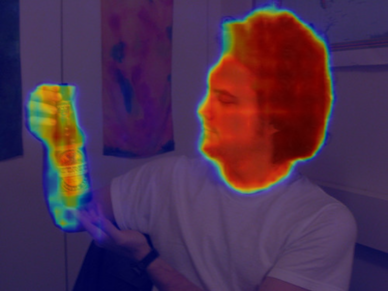}\tabularnewline
\includegraphics[width=0.4\columnwidth]{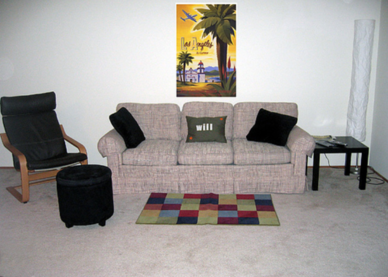} & \includegraphics[width=0.4\columnwidth]{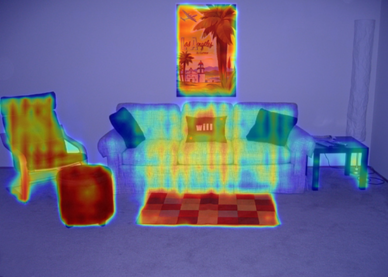} &  &  & \includegraphics[width=0.4\columnwidth]{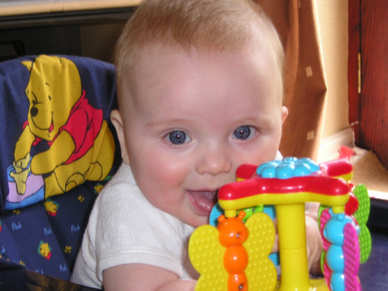} & \includegraphics[width=0.4\columnwidth]{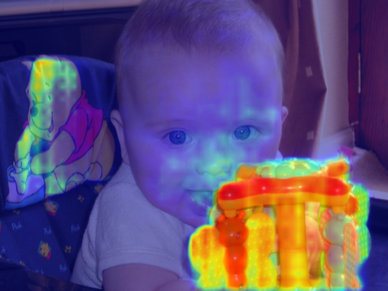}\tabularnewline
\includegraphics[width=0.4\columnwidth]{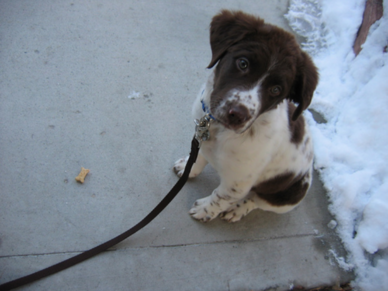} & \includegraphics[width=0.4\columnwidth]{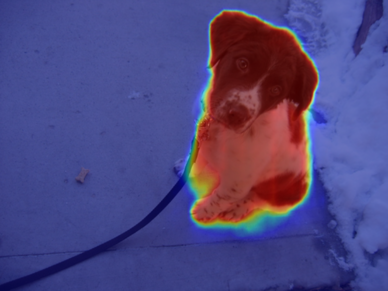} &  &  & \includegraphics[width=0.4\columnwidth]{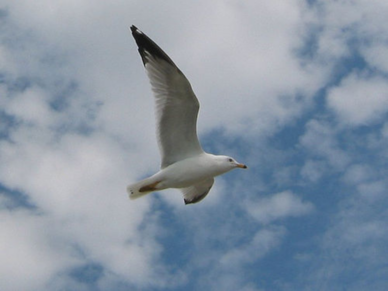} & \includegraphics[width=0.4\columnwidth]{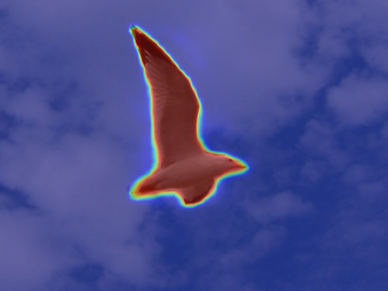}\tabularnewline
\includegraphics[width=0.4\columnwidth]{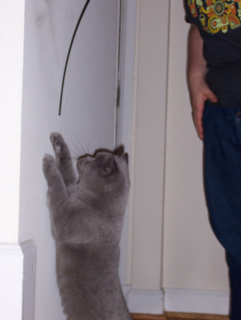} & \includegraphics[width=0.4\columnwidth]{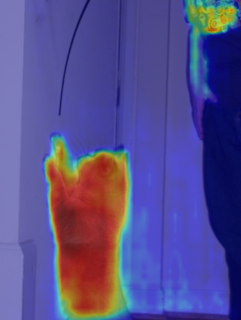} &  &  & \includegraphics[width=0.4\columnwidth]{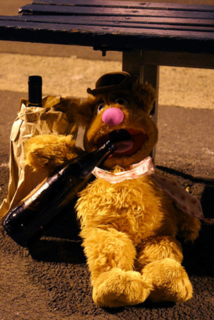} & \includegraphics[width=0.4\columnwidth]{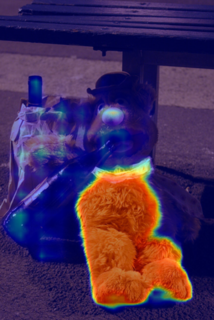}\tabularnewline
\includegraphics[width=0.4\columnwidth]{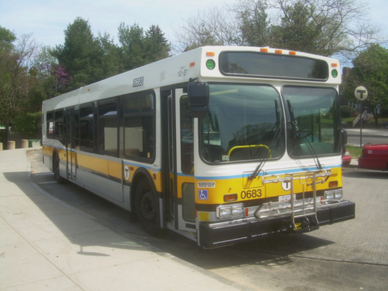} & \includegraphics[width=0.4\columnwidth]{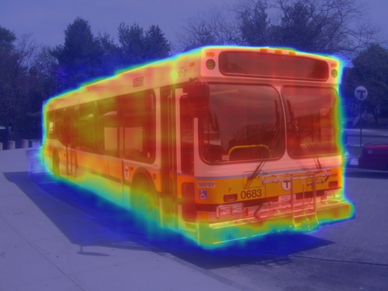} &  &  & \includegraphics[width=0.4\columnwidth]{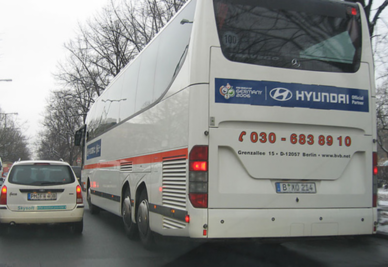} & \includegraphics[width=0.4\columnwidth]{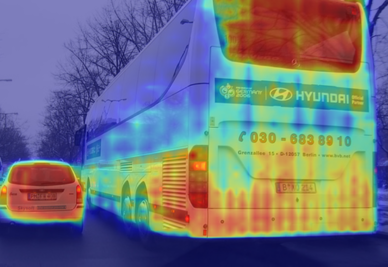}\tabularnewline
\includegraphics[width=0.4\columnwidth]{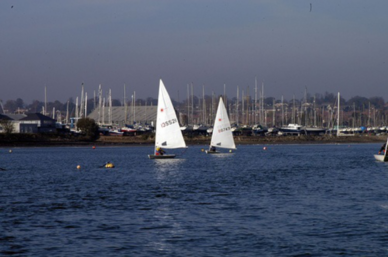} & \includegraphics[width=0.4\columnwidth]{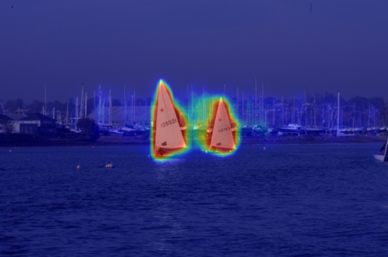} &  &  & \includegraphics[width=0.4\columnwidth]{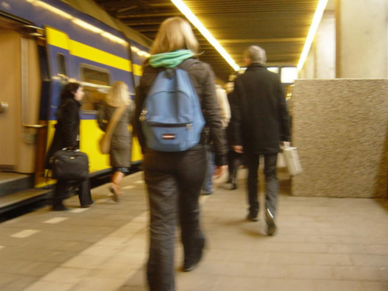} & \includegraphics[width=0.4\columnwidth]{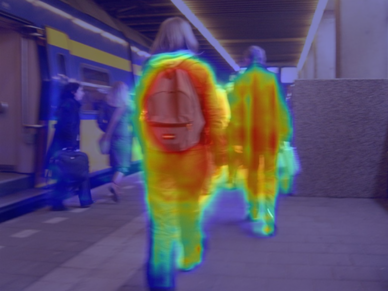}\tabularnewline
 &  & \multicolumn{1}{c}{} &  &  & \tabularnewline
\end{tabular}
\par\end{centering}
\caption{\label{fig:saliency-qualitative-1}Extension of figure \ref{fig:saliency-examples} in the main
paper. Example of saliency results on Pascal images. We note that
the saliency often fails when the central, salient objects are non-Pascal
or when the scene is cluttered. Examples are chosen at random.}
\end{figure*}

See figure \ref{fig:saliency-qualitative-1} for qualitative examples
of our saliency model on the Pascal images. We observe that the saliency
model does fail in examples usually when the central salient object
is not Pascal category, or when the scene is cluttered.

\section{\label{sec:supp-alg}$\mathcal{G}_{2}$ guide labeller algorithm}

We introduce details of the algorithm for $\mathcal{G}_{2}$ strategy
of combining seed and saliency signals (\S5.3 of the main paper).
As mentioned in the main paper, we follow five simple ideas:
\begin{enumerate}
\item We treat seeds as reliable small size point predictors of each object
instance.
\item We assume the saliency might trigger on objects that are not part
of the classes of interest.
\item If a seed touches a connected component $R_{i}^{fg},$it should take
the label of the seed.
\item If two (or more) seeds touch the same foreground component, then we
want to propagate all the seed labels inside it.
\item When in doubt, mark as ignore.
\end{enumerate}
The detailed procedure is given as follows.

We compute the set of connected components of the saliency foreground
mask with area $\geq$ $1\%$ of the image size, $\{R_{i}^{fg}\}_{i}$,
and similarly for the set of connected components of the seeds, $\{R_{j}^{s}\}_{j}$.
For each $R_{i}^{fg}$, we assign a ground truth label on it depending
on how many foreground seed categories it intersects with:
\begin{itemize}
\item 0 category: $R_{i}^{fg}$ is then either a false positive from the
saliency (e.g. salient object that is not part of the classes of interest),
or a false negative from the seeds. We don't commit to any of those
cases by marking with ``ignore'' label.
\item 1 category: $R_{i}^{fg}$ is delineating the full extent of the instance
for the seed. Put the class label from the seed.
\item $\geq2$ categories: $R_{i}^{fg}$ is a combination of instances from
multiple classes. Use dense CRF inference inside $R_{i}^{fg}$, with
unaries set by the seed(s), to assign precise pixel-wise labels in
$R_{i}^{fg}$.
\end{itemize}
After assigning pixel-wise labels on each $R_{i}^{fg}$, we perform
the following operations regarding the seed connected components $R_{j}^{s}$:
\begin{itemize}
\item When a seed $R_{j}^{s}$ intersects with some $R_{i}^{fg}$, but is
not strictly covered by $R_{i}^{fg}$, we put ``ignore'' labels
on the seed region bleeding out of $R_{i}^{fg}$, assuming that the
saliency mask provides a better delineation of the object.
\item If a seed $R_{j}^{s}$ touches two or more foreground regions, it
will propagate its label to all of them.
\item Whenever there is an isolated seed $R_{j}^{s}$ not intersecting with
any $R_{i}^{fg}$, we treat it as a reliable foreground prediction
missed by saliency, and include it in the final guide labelling. 
\end{itemize}

\begin{figure*}
\begin{centering}
\begin{tabular}{ccccccc}
Image & Seeds & Saliency & $\mathcal{G}_{0}$ & $\mathcal{G}_{1}$ & $\mathcal{G}_{2}$ & Ground truth\tabularnewline
\includegraphics[width=0.25\columnwidth]{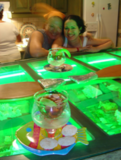} & \includegraphics[width=0.25\columnwidth]{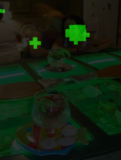} & \includegraphics[width=0.25\columnwidth]{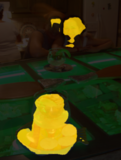} & \includegraphics[width=0.25\columnwidth]{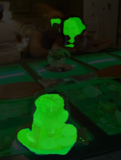} & \includegraphics[width=0.25\columnwidth]{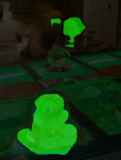} & \includegraphics[width=0.25\columnwidth]{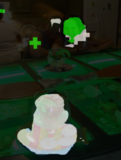} & \includegraphics[width=0.25\columnwidth]{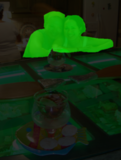}\tabularnewline
\includegraphics[width=0.25\columnwidth]{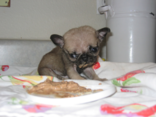} & \includegraphics[width=0.25\columnwidth]{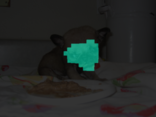} & \includegraphics[width=0.25\columnwidth]{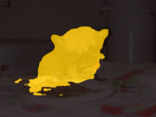} & \includegraphics[width=0.25\columnwidth]{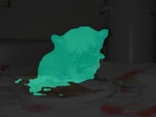} & \includegraphics[width=0.25\columnwidth]{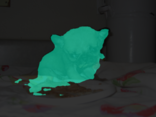} & \includegraphics[width=0.25\columnwidth]{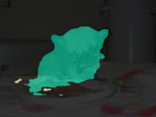} & \includegraphics[width=0.25\columnwidth]{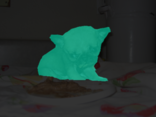}\tabularnewline
\includegraphics[width=0.25\columnwidth]{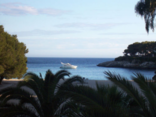} & \includegraphics[width=0.25\columnwidth]{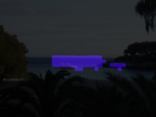} & \includegraphics[width=0.25\columnwidth]{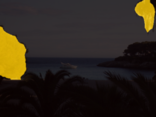} & \includegraphics[width=0.25\columnwidth]{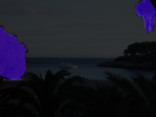} & \includegraphics[width=0.25\columnwidth]{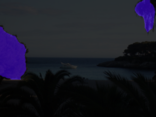} & \includegraphics[width=0.25\columnwidth]{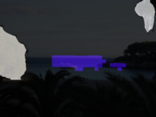} & \includegraphics[width=0.25\columnwidth]{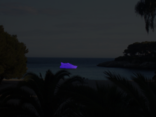}\tabularnewline
\includegraphics[width=0.25\columnwidth]{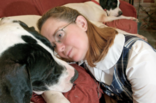} & \includegraphics[width=0.25\columnwidth]{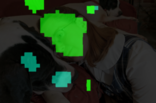} & \includegraphics[width=0.25\columnwidth]{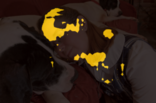} & \includegraphics[width=0.25\columnwidth]{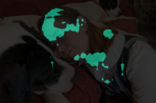} & \includegraphics[width=0.25\columnwidth]{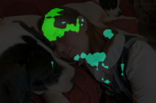} & \includegraphics[width=0.25\columnwidth]{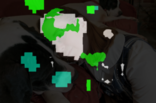} & \includegraphics[width=0.25\columnwidth]{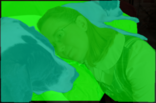}\tabularnewline
\includegraphics[width=0.25\columnwidth]{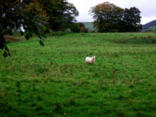} & \includegraphics[width=0.25\columnwidth]{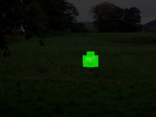} & \includegraphics[width=0.25\columnwidth]{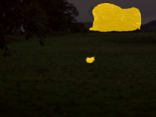} & \includegraphics[width=0.25\columnwidth]{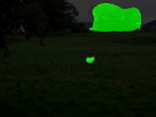} & \includegraphics[width=0.25\columnwidth]{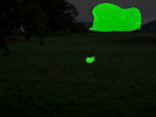} & \includegraphics[width=0.25\columnwidth]{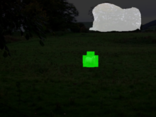} & \includegraphics[width=0.25\columnwidth]{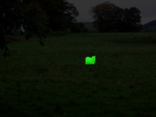}\tabularnewline
\includegraphics[width=0.25\columnwidth]{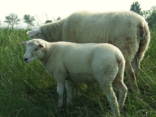} & \includegraphics[width=0.25\columnwidth]{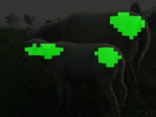} & \includegraphics[width=0.25\columnwidth]{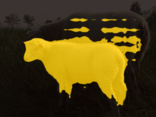} & \includegraphics[width=0.25\columnwidth]{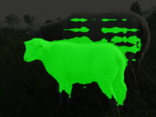} & \includegraphics[width=0.25\columnwidth]{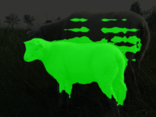} & \includegraphics[width=0.25\columnwidth]{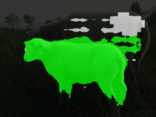} & \includegraphics[width=0.25\columnwidth]{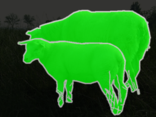}\tabularnewline
\includegraphics[width=0.25\columnwidth]{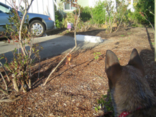} & \includegraphics[width=0.25\columnwidth]{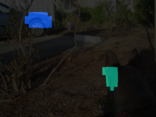} & \includegraphics[width=0.25\columnwidth]{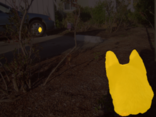} & \includegraphics[width=0.25\columnwidth]{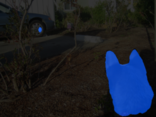} & \includegraphics[width=0.25\columnwidth]{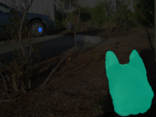} & \includegraphics[width=0.25\columnwidth]{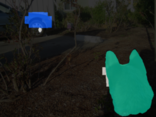} & \includegraphics[width=0.25\columnwidth]{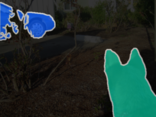}\tabularnewline
\includegraphics[width=0.25\columnwidth]{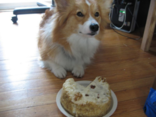} & \includegraphics[width=0.25\columnwidth]{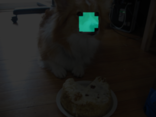} & \includegraphics[width=0.25\columnwidth]{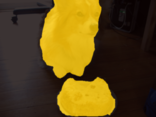} & \includegraphics[width=0.25\columnwidth]{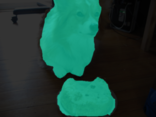} & \includegraphics[width=0.25\columnwidth]{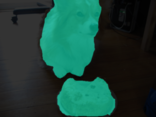} & \includegraphics[width=0.25\columnwidth]{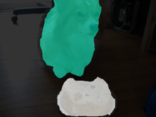} & \includegraphics[width=0.25\columnwidth]{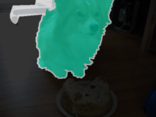}\tabularnewline
 &  &  &  &  &  & \tabularnewline
\end{tabular}
\par\end{centering}
\caption{\label{fig:guide-strategies}Extension of figure \ref{fig:combination-strategies-examples} in the main paper.
Example results for three different guide labelling strategies, $\mathcal{G}_{0}$,
$\mathcal{G}_{1}$, and $\mathcal{G}_{2}$. The image, its image labels,
seeds, and saliency map are their input. White labels indicate ``ignore''
regions. Note that $\mathcal{G}_{0}$ and $\mathcal{G}_{1}$ give
qualitatively similar results, while $\mathcal{G}_{2}$ produces much
more precise labelling by exploiting rich localisation information
from the seeds. Examples are chosen at random.}
\end{figure*}

See figure \ref{fig:guide-strategies} for the qualitative examples
of guide labelling strategies, $\mathcal{G}_{0}$, $\mathcal{G}_{1}$,
and $\mathcal{G}_{2}$. Note that $\mathcal{G}_{2}$ produces much
more precise labelling with the access to rich localisation information
from the seeds. 

\section{\label{sec:supp-convnet-training}Convnet training details}

\paragraph{Saliency}

The network is $\mathtt{DeepLab}$-v2 ResNet, and follows the training
procedure for $\mathtt{DeepLab}$-v2 ResNet in \cite{Chen2016ArxivDeeplabv2}. 

\paragraph{Segmenter}

The network is $\mathtt{DeepLab}$-v1, and is trained with stochastic
gradient descent (SGD) with minibatch size $15$, momentum $0.9$,
weight decay $5\times10^{-4}$, and base learning rate $0.001$, decreased
by the factor of $10$ at every $2\,000$ iterations. The training
stops at $8\,000$ iterations. 

\paragraph{Classifiers}

All classifiers discussed in the paper are $\mathtt{VGG}$-16 trained
with stochastic gradient descent (SGD) with minibatch size $40$,
momentum $0.9$, weight decay $5\times10^{-4}$, and base learning
rate $0.001$, decreased by the factor of $10$ at every $5\,000$
iterations. The training stops at $30\,000$ iterations. 

\section{\label{sec:supp-qualitative}Qualitative examples}

\begin{figure*}
\begin{centering}
\begin{tabular}{ccccc}
\begin{turn}{90}
Input image
\end{turn} & \includegraphics[width=0.4\columnwidth]{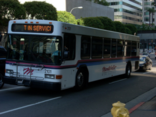} & \includegraphics[width=0.4\columnwidth]{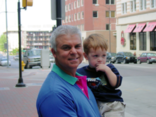} & \includegraphics[width=0.4\columnwidth]{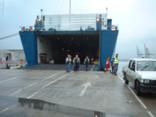} & \includegraphics[width=0.4\columnwidth]{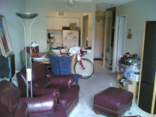}\tabularnewline
\begin{turn}{90}
Seeds\hspace*{-3em}
\end{turn} & \includegraphics[width=0.4\columnwidth]{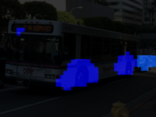} & \includegraphics[width=0.4\columnwidth]{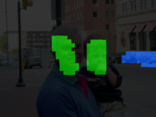} & \includegraphics[width=0.4\columnwidth]{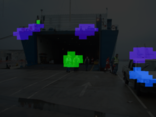} & \includegraphics[width=0.4\columnwidth]{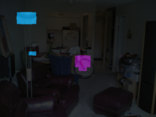}\tabularnewline
\begin{turn}{90}
Saliency\hspace*{-3em}
\end{turn} & \includegraphics[width=0.4\columnwidth]{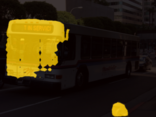} & \includegraphics[width=0.4\columnwidth]{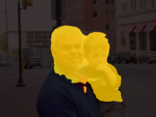} & \includegraphics[width=0.4\columnwidth]{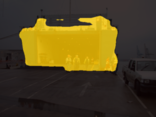} & \includegraphics[width=0.4\columnwidth]{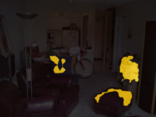}\tabularnewline
\begin{turn}{90}
\begin{tabular}{c}
$\mathcal{G}_{2}$\tabularnewline
{\small{}(Seeds+Saliency)}\tabularnewline
\end{tabular}
\end{turn} & \includegraphics[width=0.4\columnwidth]{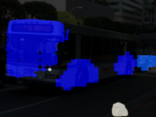} & \includegraphics[width=0.4\columnwidth]{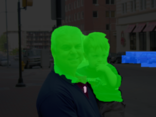} & \includegraphics[width=0.4\columnwidth]{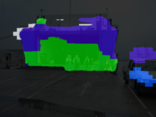} & \includegraphics[width=0.4\columnwidth]{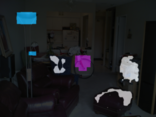}\tabularnewline
\begin{turn}{90}
\begin{tabular}{c}
Segmenter\tabularnewline
output\tabularnewline
\end{tabular}
\end{turn} & \includegraphics[width=0.4\columnwidth]{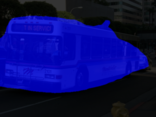} & \includegraphics[width=0.4\columnwidth]{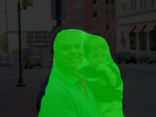} & \includegraphics[width=0.4\columnwidth]{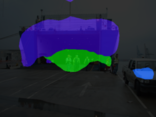} & \includegraphics[width=0.4\columnwidth]{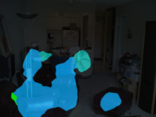}\tabularnewline
\begin{turn}{90}
+CRF\hspace*{-3em}
\end{turn} & \includegraphics[width=0.4\columnwidth]{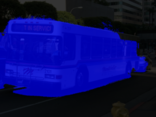} & \includegraphics[width=0.4\columnwidth]{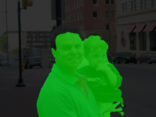} & \includegraphics[width=0.4\columnwidth]{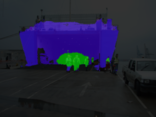} & \includegraphics[width=0.4\columnwidth]{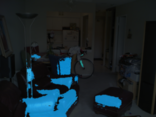}\tabularnewline
\begin{turn}{90}
Ground truth
\end{turn} & \includegraphics[width=0.4\columnwidth]{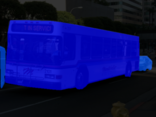} & \includegraphics[width=0.4\columnwidth]{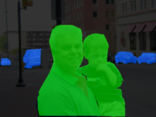} & \includegraphics[width=0.4\columnwidth]{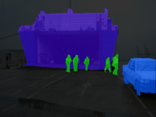} & \includegraphics[width=0.4\columnwidth]{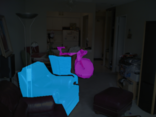}\tabularnewline
 &  &  &  & \tabularnewline
\end{tabular}
\par\end{centering}
\caption{\label{fig:qualitative-examples-1}Extension of figure \ref{fig:qualitative-examples} in the main
paper. Qualitative examples of the different stages of the Guided
Segmentation system on the training images. White labels are ``ignore''
regions. Seeds have high precision and low recall; combined with saliency
foreground mask using $\mathcal{G}_{2}$ guide labeller, object extents
are recovered. The generated guide labelling can still be noisy; however,
the segmenter convnet can average out the noise to produce more precise
predictions. CRF post-processing further refines the predictions.}
\end{figure*}

\begin{figure*}
\begin{centering}
\begin{tabular}{cccccc}
\begin{turn}{90}
Input image
\end{turn} & \includegraphics[height=0.35\columnwidth]{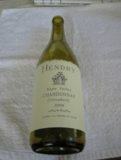} & \includegraphics[height=0.35\columnwidth]{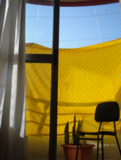} & \includegraphics[height=0.35\columnwidth]{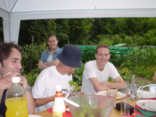} & \includegraphics[height=0.35\columnwidth]{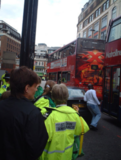} & \includegraphics[height=0.35\columnwidth]{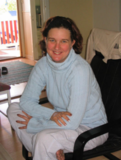}\tabularnewline
\begin{turn}{90}
Seeds\hspace*{-3em}
\end{turn} & \includegraphics[height=0.35\columnwidth]{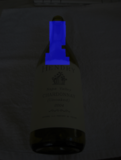} & \includegraphics[height=0.35\columnwidth]{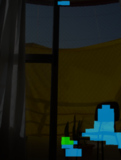} & \includegraphics[height=0.35\columnwidth]{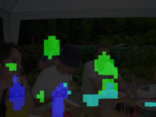} & \includegraphics[height=0.35\columnwidth]{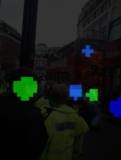} & \includegraphics[height=0.35\columnwidth]{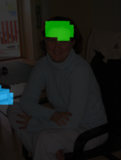}\tabularnewline
\begin{turn}{90}
Saliency\hspace*{-3em}
\end{turn} & \includegraphics[height=0.35\columnwidth]{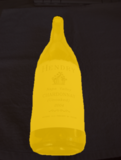} & \includegraphics[height=0.35\columnwidth]{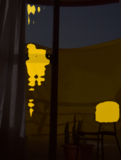} & \includegraphics[height=0.35\columnwidth]{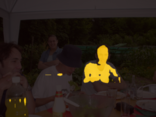} & \includegraphics[height=0.35\columnwidth]{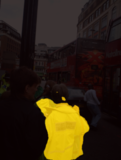} & \includegraphics[height=0.35\columnwidth]{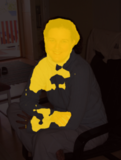}\tabularnewline
\begin{turn}{90}
\begin{tabular}{c}
$\mathcal{G}_{2}$\tabularnewline
{\small{}(Seeds+Saliency)}\tabularnewline
\end{tabular}
\end{turn} & \includegraphics[height=0.35\columnwidth]{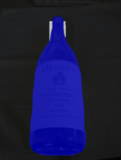} & \includegraphics[height=0.35\columnwidth]{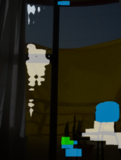} & \includegraphics[height=0.35\columnwidth]{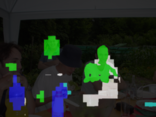} & \includegraphics[height=0.35\columnwidth]{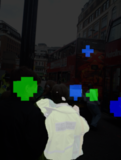} & \includegraphics[height=0.35\columnwidth]{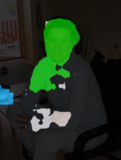}\tabularnewline
\begin{turn}{90}
\begin{tabular}{c}
Segmenter\tabularnewline
output\tabularnewline
\end{tabular}
\end{turn} & \includegraphics[height=0.35\columnwidth]{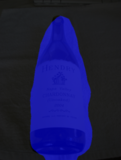} & \includegraphics[height=0.35\columnwidth]{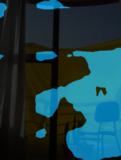} & \includegraphics[height=0.35\columnwidth]{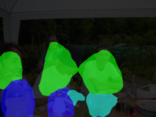} & \includegraphics[height=0.35\columnwidth]{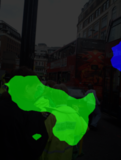} & \includegraphics[height=0.35\columnwidth]{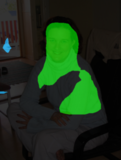}\tabularnewline
\begin{turn}{90}
+CRF\hspace*{-3em}
\end{turn} & \includegraphics[height=0.35\columnwidth]{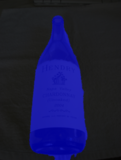} & \includegraphics[height=0.35\columnwidth]{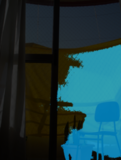} & \includegraphics[height=0.35\columnwidth]{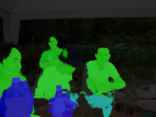} & \includegraphics[height=0.35\columnwidth]{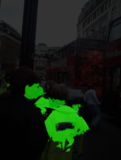} & \includegraphics[height=0.35\columnwidth]{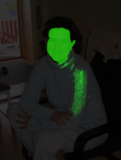}\tabularnewline
\begin{turn}{90}
Ground truth
\end{turn} & \includegraphics[height=0.35\columnwidth]{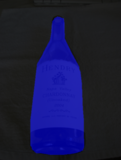} & \includegraphics[height=0.35\columnwidth]{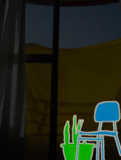} & \includegraphics[height=0.35\columnwidth]{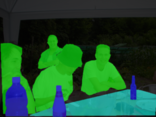} & \includegraphics[height=0.35\columnwidth]{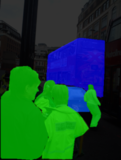} & \includegraphics[height=0.35\columnwidth]{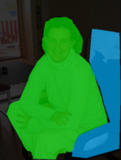}\tabularnewline
 &  &  &  &  & \tabularnewline
\end{tabular}
\par\end{centering}
\caption{\label{fig:qualitative-examples-2}Extension of figure \ref{fig:qualitative-examples} in the main
paper. More qualitative examples of the different stages of the Guided
Segmentation system on the training images. White labels are ``ignore''
regions. Seeds have high precision and low recall; combined with saliency
foreground mask using $\mathcal{G}_{2}$ guide labeller, object extents
are recovered. The generated guide labelling can still be noisy; however,
the segmenter convnet can average out the noise to produce more precise
predictions. CRF post-processing further refines the predictions.}
\end{figure*}

See figure \ref{fig:qualitative-examples-1} and \ref{fig:qualitative-examples-2}
for more qualitative examples of the seeds, saliency, $\mathcal{G}_{2}$
guide labeller output, and Guided Segmentation trained results on
the training set. Seeds have high precision and low recall. The saliency
foreground mask gives a pixel-wise class-agnostic object extent information.
$\mathcal{G}_{2}$ guide labeller combines both sources to generate
an accurate class-wise guide labelling. The generated guide labelling
can still be noisy especially if the quality of the saliency mask
is low. However, the segmenter convnet averages out the noisy supervision
to produce more precise predictions. CRF post-processing further refines
the predictions.

\newpage



\begin{table*}
\caption{\label{tab:GAP-networks}Detailed architecture of the GAP networks.
Triplets denote (channel, height, width) for the input and output
data, and (output channel dim, kernel height, kernel width) for the
layer parameters. $\mathtt{st}=$stride and $\mathtt{dil}=$width
of dilated convolution, with default values $1$ for both, unless
otherwise stated.}

\begin{centering}
\begin{tabular}{cccccc}
 &  &  &  &  & \tabularnewline
Layers & $\mathtt{VGG}\text{-}\mathtt{16}$ & $\mathtt{GAP}\text{-}\mathtt{LowRes}$ &  $\mathtt{GAP}\text{-}\mathtt{HighRes}$ &  $\mathtt{GAP}\text{-}\mathtt{ROI}$ &  $\mathtt{GAP}\text{-}\mathtt{DeepLab}$\tabularnewline
 & \cite{Simonyan2015Iclr} &  \cite{zhou2015cnnlocalization} & \cite{kolesnikov2016seed} &  & \cite{Chen2016ArxivDeeplabv2}\tabularnewline
\hline 
$\mathtt{input}$ (C, H, W) & $(3,224,224)$ & $(3,321,321)$ & $(3,321,321)$ & $(3,321,321)$ & $(3,321,321)$\tabularnewline
\hline 
$2\times\mathtt{conv1}$ & $(64,3,3)$ & Same as & Same as & Same as & Same as\tabularnewline
 & $\mathtt{pad}=1$ & $\mathtt{VGG}\text{-}\mathtt{16}$ & $\mathtt{VGG}\text{-}\mathtt{16}$ & $\mathtt{VGG}\text{-}\mathtt{16}$ & $\mathtt{VGG}\text{-}\mathtt{16}$\tabularnewline
\hline 
$\mathtt{pool1}$ & $(\text{-},2,2)$ & Same as & Same as & Same as & $(\text{-},3,3)$\tabularnewline
 & $\mathtt{pad}=0$ & $\mathtt{VGG}\text{-}\mathtt{16}$ & $\mathtt{VGG}\text{-}\mathtt{16}$ & $\mathtt{VGG}\text{-}\mathtt{16}$ & $\mathtt{st}=2$, $\mathtt{pad}=1$\tabularnewline
\hline 
$2\times\mathtt{conv2}$ & $(128,3,3)$ & Same as & Same as & Same as & Same as\tabularnewline
 & $\mathtt{pad}=1$ & $\mathtt{VGG}\text{-}\mathtt{16}$ & $\mathtt{VGG}\text{-}\mathtt{16}$ & $\mathtt{VGG}\text{-}\mathtt{16}$ & $\mathtt{VGG}\text{-}\mathtt{16}$\tabularnewline
\hline 
$\mathtt{pool2}$ & $(\text{-},2,2)$ & Same as & Same as & Same as & $(\text{-},3,3)$\tabularnewline
 & $\mathtt{pad}=0$ & $\mathtt{VGG}\text{-}\mathtt{16}$ & $\mathtt{VGG}\text{-}\mathtt{16}$ & $\mathtt{VGG}\text{-}\mathtt{16}$ & $\mathtt{st}=2$, $\mathtt{pad}=1$\tabularnewline
\hline 
$3\times\mathtt{conv3}$ & $(256,3,3)$ & Same as & Same as & Same as & Same as\tabularnewline
 & $\mathtt{pad}=1$ & $\mathtt{VGG}\text{-}\mathtt{16}$ & $\mathtt{VGG}\text{-}\mathtt{16}$ & $\mathtt{VGG}\text{-}\mathtt{16}$ & $\mathtt{VGG}\text{-}\mathtt{16}$\tabularnewline
\hline 
$\mathtt{pool3}$ & $(\text{-},2,2)$ & Same as & Same as & Same as & $(\text{-},3,3)$\tabularnewline
 & $\mathtt{pad}=0$ & $\mathtt{VGG}\text{-}\mathtt{16}$ & $\mathtt{VGG}\text{-}\mathtt{16}$ & $\mathtt{VGG}\text{-}\mathtt{16}$ & $\mathtt{st}=1$, $\mathtt{pad}=1$\tabularnewline
\hline 
$3\times\mathtt{conv4}$ & $(512,3,3)$ & Same as & Same as & Same as & Same as\tabularnewline
 & $\mathtt{pad}=1$ & $\mathtt{VGG}\text{-}\mathtt{16}$ & $\mathtt{VGG}\text{-}\mathtt{16}$ & $\mathtt{VGG}\text{-}\mathtt{16}$ & $\mathtt{VGG}\text{-}\mathtt{16}$\tabularnewline
\hline 
$\mathtt{pool4}$ & $(\text{-},2,2)$ & Same as & None & None & $(\text{-},3,3)$\tabularnewline
 & $\mathtt{pad}=0$ & $\mathtt{VGG}\text{-}\mathtt{16}$ &  &  & $\mathtt{st}=1$, $\mathtt{pad}=1$\tabularnewline
\hline 
$3\times\mathtt{conv5}$ & $(512,3,3)$ & Same as & Same as & Same as & $(512,3,3)$\tabularnewline
 & $\mathtt{pad}=1$ & $\mathtt{VGG}\text{-}\mathtt{16}$ & $\mathtt{VGG}\text{-}\mathtt{16}$ & $\mathtt{VGG}\text{-}\mathtt{16}$ & $\mathtt{dil}=2$, $\mathtt{pad}=2$\tabularnewline
\hline 
$\mathtt{pool5}$ & $(\text{-},2,2)$ & None & None & ROI-pool & None\tabularnewline
 & $\mathtt{pad}=0$ &  &  & $3\times3$ windows & \tabularnewline
\hline 
$\mathtt{fc6}$ & $(4096,7,7)$ & $(1024,3,3)$ & $(1024,3,3)$ & $(1024,1,1)$ & $(1024,3,3)$\tabularnewline
 & $\mathtt{pad}=0$ & $\mathtt{pad}=1$ & $\mathtt{pad}=1$ & $\mathtt{pad}=0$ & $\mathtt{dil}=12$, $\mathtt{pad}=12$\tabularnewline
\hline 
$\mathtt{fc7}$ & $(4096,1,1)$ & None & $(1024,3,3)$ & $(1024,1,1)$ & $(1024,1,1)$\tabularnewline
 & $\mathtt{pad}=0$ &  & $\mathtt{pad}=1$ & $\mathtt{pad}=0$ & $\mathtt{pad}=0$\tabularnewline
\hline 
$\mathtt{GAP}$ & None & $\mathtt{GAP}$ & $\mathtt{GAP}$ & None (used after $\mathtt{fc}8$) & $\mathtt{GAP}$\tabularnewline
\hline 
$\mathtt{fc8}$ & $(20,1,1)$ & Same as & Same as & Same as & Same as\tabularnewline
 & $\mathtt{pad}=0$ & $\mathtt{VGG}\text{-}\mathtt{16}$ & $\mathtt{VGG}\text{-}\mathtt{16}$ & $\mathtt{VGG}\text{-}\mathtt{16}$ & $\mathtt{VGG}\text{-}\mathtt{16}$\tabularnewline
\hline 
$\mathtt{output}$ heatmap & $(20,1,1)$ & $(20,21,21)$ & $(20,41,41)$ & $(20,41,41)$ & $(20,41,41)$\tabularnewline
\hline 
 &  &  &  &  & \tabularnewline
\end{tabular}
\par\end{centering}
\end{table*}

\newpage

\end{document}